\documentclass[journal]{IEEEtran}

\usepackage{cite}
   \usepackage{amsmath}
   \usepackage{mathptmx}

\usepackage[]{algorithm2e}
\usepackage{algpseudocode}
\usepackage{listings}
\usepackage{graphicx}
\newtheorem{defn}{Definition}
\usepackage{soul, color}\setuldepth{article}
\usepackage{amsmath}
\usepackage[dvipsnames]{xcolor}
\usepackage{tabularx}
\ifCLASSINFOpdf
\else
\fi

\begin{document}
%
\title{Color-Emotion Associations in Art: Fuzzy Approach}


\author{\IEEEauthorblockN{Muragul Muratbekova \IEEEauthorrefmark{1},
Pakizar Shamoi\IEEEauthorrefmark{2}}
\\
\IEEEauthorblockA{School of Information Technology and Engineering \\
Kazakh-British Technical University\\
Almaty, Kazakhstan\\
Email: \IEEEauthorrefmark{1}mu\_muratbekova@kbtu.kz,
\IEEEauthorrefmark{2}p.shamoi@kbtu.kz,
}
}


%


\maketitle



%
\IEEEpeerreviewmaketitle

\begin{abstract}

Art objects can evoke certain emotions. Color is a fundamental element of visual art and plays a significant role in how art is perceived. This paper introduces a novel approach to classifying emotions in art using Fuzzy Sets. We employ a fuzzy approach because it aligns well with human judgments' imprecise and subjective nature. Extensive fuzzy colors (n=120) and a broad emotional spectrum (n=10) allow for a more human-consistent and context-aware exploration of emotions inherent in paintings. First, we introduce the fuzzy color representation model. Then, at the fuzzification stage, we process the Wiki Art Dataset of paintings tagged with emotions, extracting fuzzy dominant colors linked to specific emotions. This results in fuzzy color distributions for ten emotions. Finally, we convert them back to a crisp domain, obtaining a knowledge base of color-emotion associations in primary colors. Our findings reveal strong associations between specific emotions and colors;  for instance, gratitude strongly correlates with green, brown, and orange. Other noteworthy associations include brown and anger, orange with shame, yellow with happiness, and gray with fear. Using these associations and Jaccard similarity, we can find the emotions in the arbitrary untagged image. We conducted a 2AFC experiment involving human subjects to evaluate the proposed method. The average hit rate of 0.77 indicates a significant correlation between the method's predictions and human perception. The proposed method is simple to adapt to art painting retrieval systems. The study contributes to the theoretical understanding of color-emotion associations in art, offering valuable insights for various practical applications besides art, like marketing, design, and psychology.

\end{abstract}

\begin{IEEEkeywords}
Fuzzy sets, emotions in art, color palette, classification, color-emotion model, art image analysis, color perception, image processing emotion detection.
\end{IEEEkeywords}

\maketitle

\section{Introduction}




Nowadays, an increasing amount of affective information is distributed worldwide in all formats \cite{Cambria2016}. The elements that contribute to the sensory experience of audio and visual content include acoustic quality, facial expressions and gestures, background music, ambient noise, color schemes, and image filters. It is crucial to assess incoming data objectively, including emotional reactions. \cite{Ivanova2010, Ou2004}.

Color is a fundamental element of visual art and plays a big role in the emotional responses induced by the art image. It is important to classify art not just in terms of age or style but also in terms of the emotions they create in a viewer. That can facilitate the identification of art expressing similar emotions and has a potential application in context-based image retrieval systems.

Emotion detection aims to determine the presence and intensity of emotion as close as possible to human perception. We need a machine that can objectively perceive objects despite subjective color perception. \cite{Hibadullah2015}.

The relationship between color and emotion has been studied in various areas, ranging from psychology to art theory. 

\begin{figure*}[ht]
  \includegraphics[width=\textwidth]{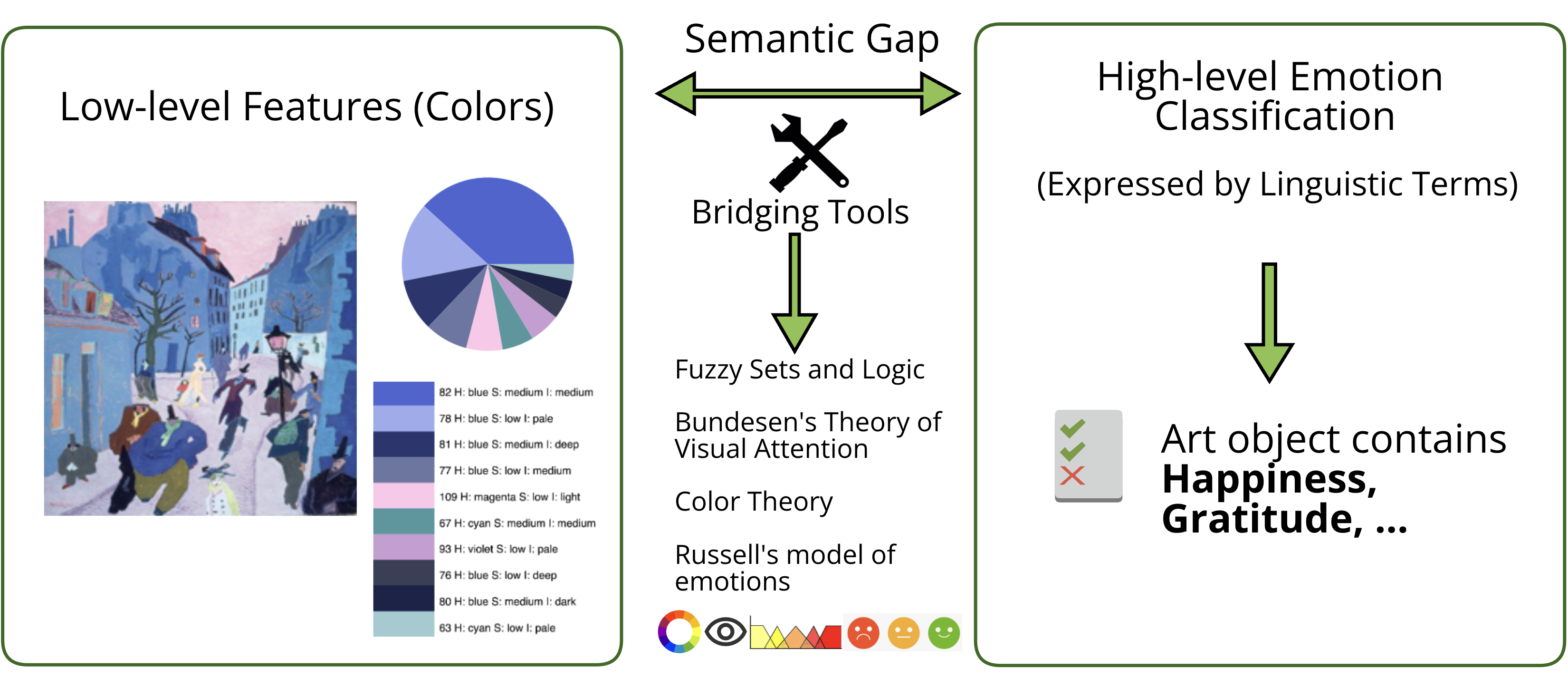}
\caption{Bridging the semantic gap between low-level features in art objects and high-level semantic concepts of emotions. Lyonel Feininger "Carnival in Arcueil" painting.}
\label{gap}
\end{figure*}

The colors we see can greatly impact our emotions. Researchers such as J.H. Xin have found that the lightness and chroma of colors are more influential than the hue itself \cite{Xin2004}. Other studies, including one conducted by Banu Manav \cite{Manav2007}, have also shown that our emotional response to colors can vary based on their lightness and saturation. In a related experiment, the researchers in \cite{Wang2008} explored how changes in valence and arousal affected emotional response. They were able to accurately map 4 different emotions to 5 parameters: lightness, chroma, hue, valence, and arousal, with an accuracy rate of 80\%.

Currently, there are several gaps in research on this topic. Technically, there is a limitation in recognizing and comprehending pictures. Aesthetically, there is no apparent correlation between the information conveyed by a picture and the response to it.

The interaction between different emotions is still poorly understood, and the studies conducted so far have been limited to a handful of emotions (usually 4 to 6). This is because emotions are difficult to represent using a specific metric. Moreover, it is important to shift the focus from the content of paintings to their forms. However, it seems impossible to achieve universal calculations in this regard.

It is crucial to acknowledge that various factors influence perception and should be evaluated in different contexts. In this paper, our focus is on art and we present a technique to extract the emotional color palette from art objects by using fuzzy sets. This method is particularly useful as color ambiguity and perception are both subjective and inaccurate by nature. Traditional color-emotion studies have primarily relied on basic color categories, which may impact emotional experiences evoked by art.

The current paper proposes an emotion classification approach for art objects using Fuzzy Sets and Logic Approach (see Fig. \ref{gap}). It matches colors in the art domain to the spectrum of human emotions. The fuzzy approach is well-suited for color and emotion detection. How people perceive color and emotion is inherently subjective and differs from person to person. The fuzzy approach can effectively represent this imprecision in the perception of color-emotion associations in a way that classical binary logic might struggle to capture.

As it can be seen in Fig. \ref{gap}, we aim to bridge the gap between low-level features, specifically colors, and high-level emotion classifications (e.g., happy, fearful, etc.), utilizing a comprehensive set of bridging tools, including fuzzy sets and logic, the Russell emotion model, Bundesen's theory of visual attention, and color theory.


The contributions of this paper are the following:
 \begin{itemize}
     \item \textit{Enhanced emotional spectrum.}We consider a broad range of emotions (N=10) with contextual relevance in art.
     \item \textit{Increased granularity.} Basic colors provide a general understanding of color-emotion relationships, but using a wider array of fuzzy colors (N=120), our model imitates human perception of colors. This enables a richer exploration of emotions and their connections to specific color variations and combinations.
     \item \textit{Rich color expressiveness of emotions.} We employ 10-color palettes to depict the complexity inherent in art images better, but we also consider color proportion. Most studies concentrate on emotions within two or three color palettes.
     
      \item \textit{Proof of concept.} we also provided a method for automatically tagging the art objects for Art Gallery Systems for emotion-based image retrieval.
 \end{itemize}

The paper has the following structure. Section I is this Introduction. Section II presents a review of prior research on color-emotion associations. In Section III, various methods we use in this study, including fuzzy sets and logic and theory of visual attention, are detailed, along with explanations of data collection and proposed approach details. The subsequent section, Section IV, showcases the sample application and experimental results. Next, the Discussion is presented in Section V. Finally, Section VI concludes the paper and offers recommendations for future enhancements to the methodology. 

\section{Related Work}
\subsection{Color-Emotion Association Studies}

\begin{figure}
\centerline{\includegraphics[scale=0.3]{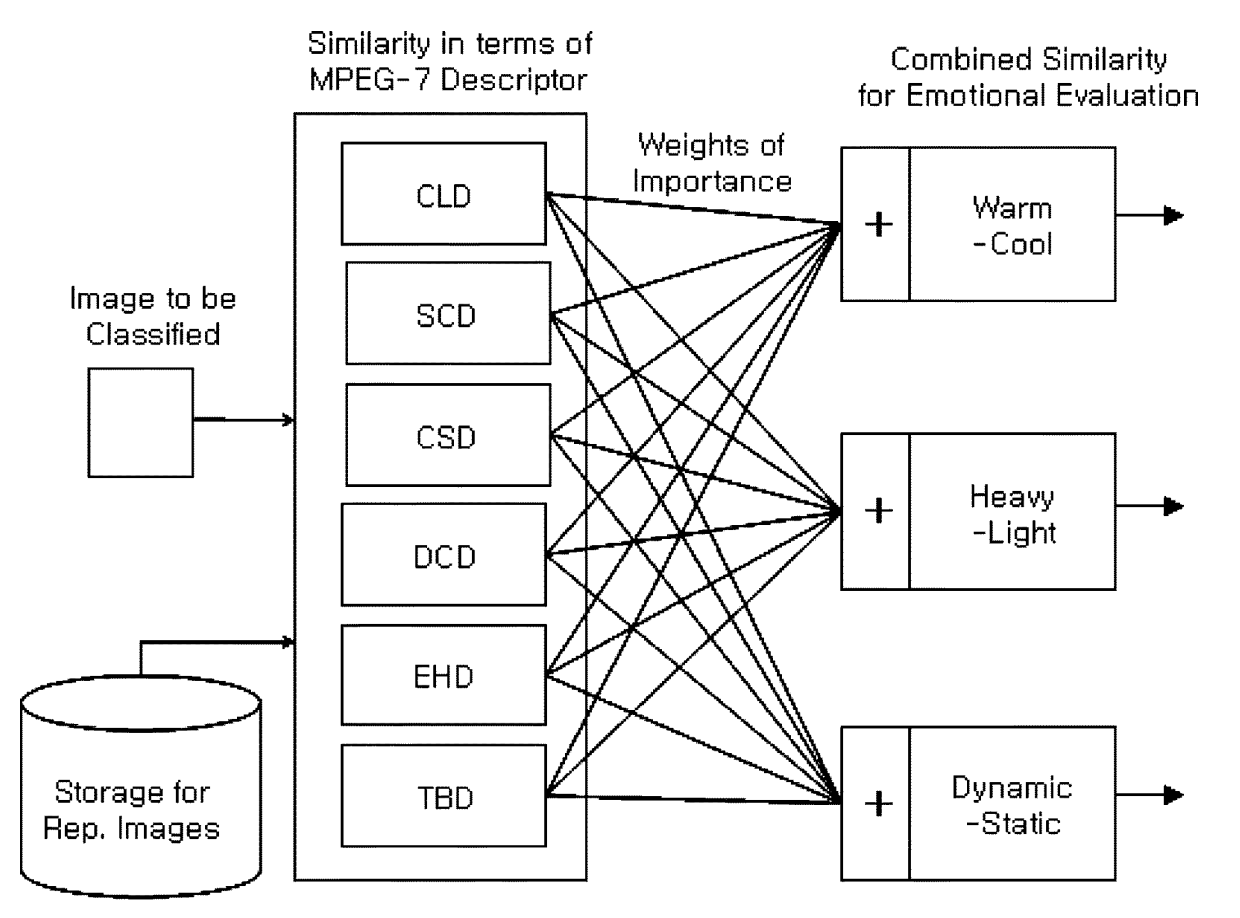}}
\caption{Lee J. Emotional classification of color images. \cite{Lee2011}}
\label{fig:Lee2011}
\end{figure}

This section presents an overview of literature related to emotion detection in Art. Several studies have examined the color-emotion associations in various contexts, including art. Prior work can be divided into three main directions: context dependency investigation, machine learning algorithms, and fuzzy approach.

The earliest studies were highly interested in the human perception subjectivity level. Color perception depends on factors such as sex, nationality, culture, etc. Authors of \cite{Gao2007} decided to find out to what extent human perception is influenced by cultural background. They compared the emotional responses of people in 7 different regions of Europe and Asia and found out that regardless of their national differences, people of different cultures reacted to colors quite similarly. So, the cultural context has very limited influence on human perception. However, the human response to color is mainly influenced by lightness, chroma, and hue. 

Several studies have examined that the aesthetics of art objects depends on their texture, cultural context, and the viewer's emotional state \cite{Xin2016} \cite{Joshi2011}. The work proposed in \cite{Joshi2011} uses Kansei engineering to translate feelings into specific parameters for frequently encountered subjects in aesthetics and emotion. 
The paper by \cite{Solli2011} conducted several psychological experiments to evaluate how the perception differs for single-color and multiple-color paintings. They used different scale types and concluded that the emotional response to color is more universal. 
Two more critical factors influence the outcome of emotional perception: order and learning. The initial factor involves components like symmetry, rhythm, repetition, and contrast. The second factor, learning, highlights that individuals tend to perceive something as aesthetically pleasing when it aligns with their previous experiences \cite{Hoenig2018}. 

\begin{figure}[tb]
\centerline{\includegraphics[scale=0.3]{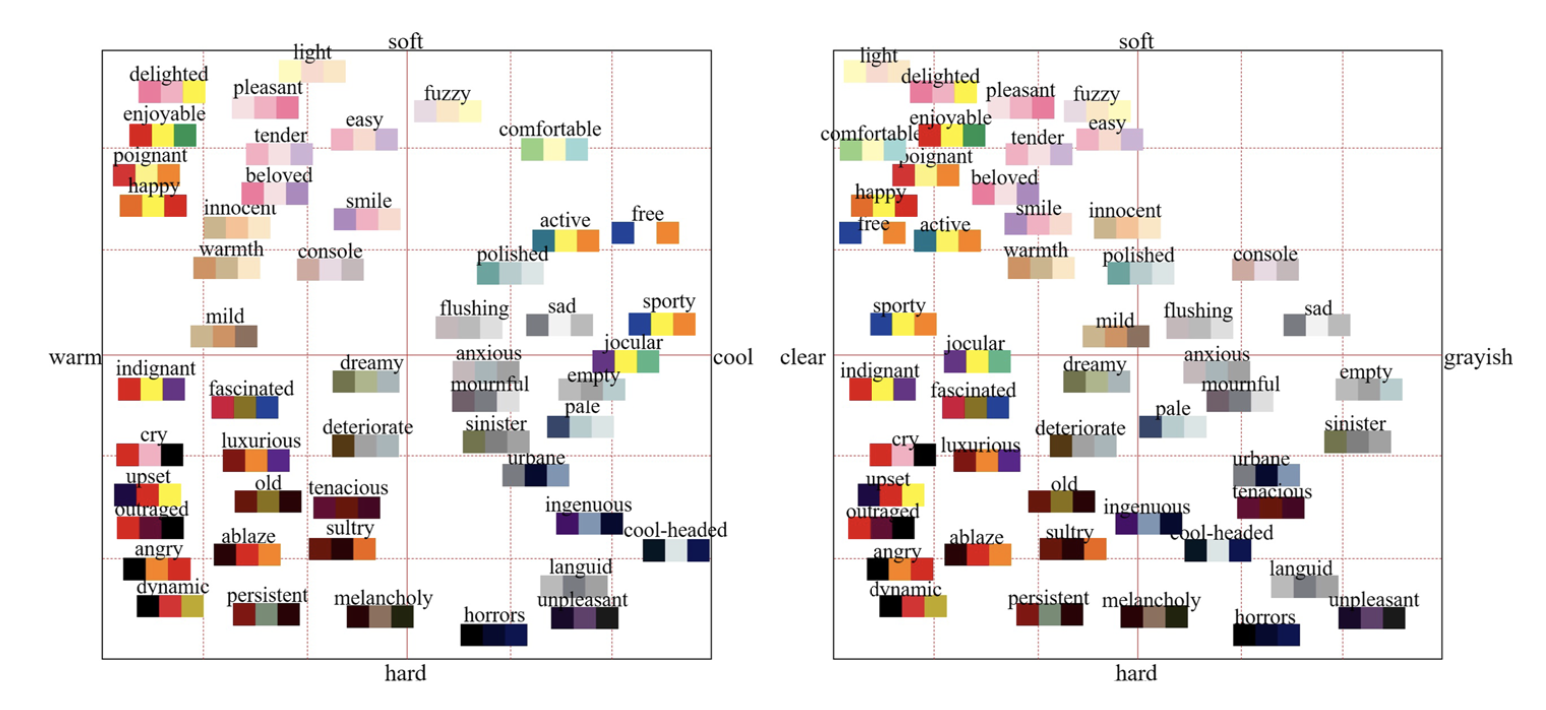}}
\caption{Kang's study color image scale \cite{Kang2018}. }
\label{fig:Kang2018}
\end{figure}

From the perspective of the proposed approach, previous works can be divided into machine learning algorithms and fuzzy logic approaches. 

Advanced search engines are mostly based on text, keywords, or using face recognition algorithms. \cite{Solli2008}, \cite{Lenz2010} proposed to use a color as a new metric for image retrieval and classification. The work of \cite{Solli2009} extended emotion recognition methods by adding the relationship between neighboring emotions. Their algorithm considers that emotion and transitions between emotion regions may influence human perception. 

The study \cite{Lee2011} provided several classification methods for emotions, such as Decision trees, support vector machines, and Naive Bayes. Additionally, it stressed the importance of selecting emotional adjectives based on four criteria: universality, distinctiveness, utility, and comprehensiveness (see Fig. \ref{fig:Lee2011}).
From the viewpoint of affective computing, it is currently understood that emotional states can only be described explicitly and not implicitly. The primary objective of affective computing is to identify and classify emotions accordingly. Several approaches have been reported in the literature: knowledge base, statistical methods, and hybrid \cite{Cambria2016}. Some of them lack semantic strength.

In their theory, Kobayashi created a graph to identify the combination of most similar colors and determine the leading emotion from it \cite{Kobayashi2009}. The work proposed in \cite{Kang2018} uses Kobayashi's color theories to extract complex emotions from paintings by adding a grayish/clear factor for improved accuracy. However, with many different emotions present in the picture, the probability of error increases. Figure \ref{fig:Kang2018} illustrates this approach.

\begin{figure}[tb]
    \centering
    \includegraphics[scale=0.3]{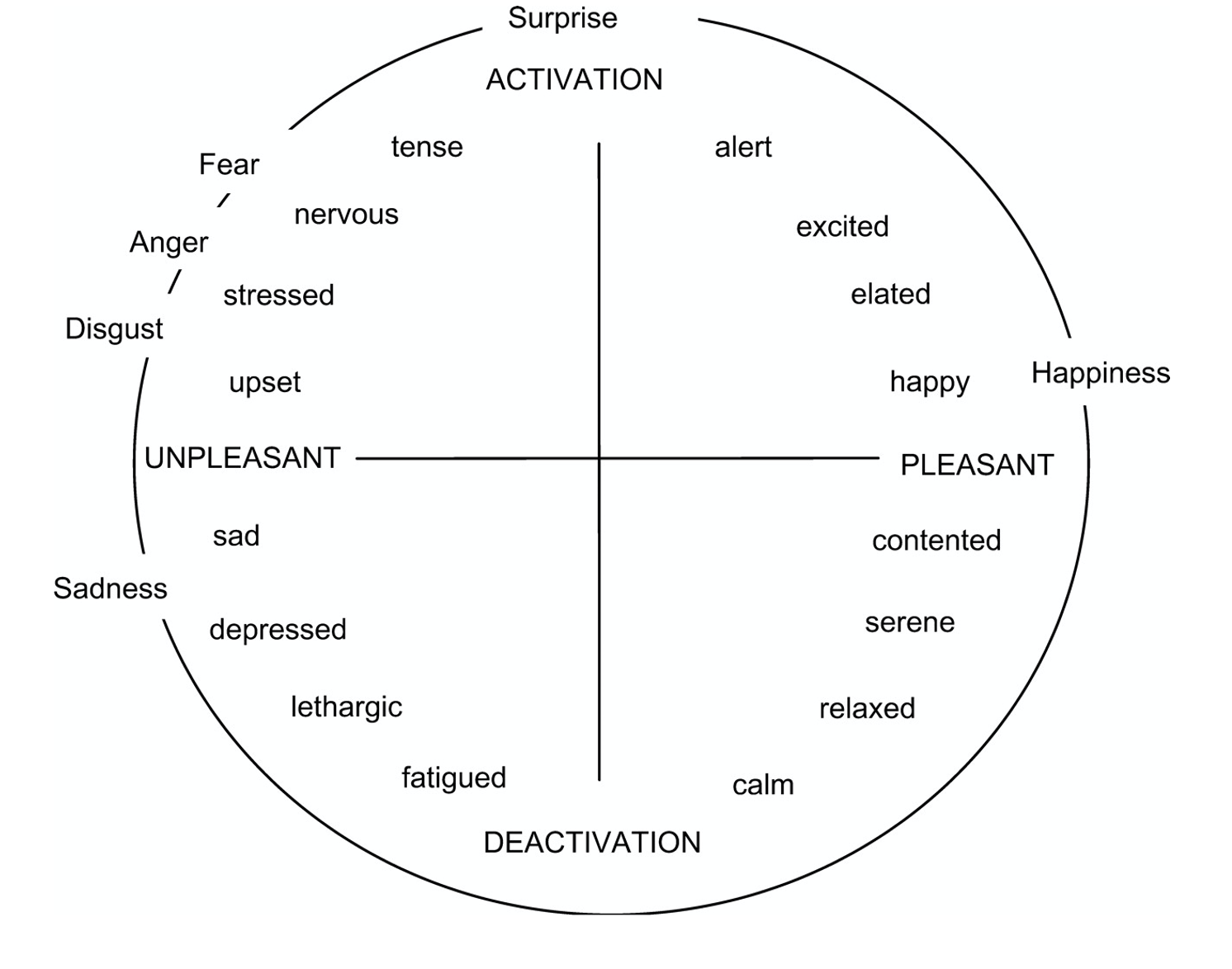}
    \caption{Russell emotional model. \cite{Russell1980}.}
    \label{fig:Russell}
\end{figure}

The authors of the other study designed a machine-learning algorithm for emotion classification in images, utilizing color information and categorizing emotions into three distinct classes: negative, positive, and neutral \cite{access_em}.
Another study proposed the L-EEP method, utilizing Culture Technology, to automatically extract eight emotions from Iranian-Islamic paintings through color analysis and the automated Luscher test \cite{access_em2}.
Recent work proposed a  network for recognizing emotions in images, utilizing a weakly supervised approach to learn emotion intensity \cite{access_em3}.

Fuzzy logic has been used to address these concerns, specifically targeting diverse contexts such as gender, age, ethnic group, and environmental factors when computing the outcome of an emotional assessment. Several works stressed the importance of the fuzzy sets approach to imitate human perception of high-level color concepts:


\begin{itemize}
    \item  Researchers introduced fuzzy color spaces defined using RGB and Euclidean distance, with examples from the well-known ISCC-NBS color naming system \cite{fuzzycolor}. 
    \item The other works introduced fuzzy granular colors constructed by consolidating fuzzy colors that share semantic relationships within a specific color category \cite{fuzzycolor2}, a method centered on images for constructing fuzzy color spaces within defined contexts \cite{fuzzycolor_3}.
    \item Nachtegael introduced a formula for calculating similarity of fuzzy color histograms.  \cite{Nachtegael2007}.
    \item In the other work, the connection between colors and emotions was investigated by employing fuzzy feature vectors derived from fuzzy color histograms and emotion vectors obtained through surveys \cite{Hibadullah2015}.
    \item Another model employed fuzzy logic and L*a*b* color space for emotional semantic queries. The generation of semantic terms was facilitated by applying a fuzzy clustering algorithm \cite{1527744}.
    \item Recent research by \cite{Shamoi2022} used fuzzy logic to explore the universality of color palettes concerning specific associations. The study revealed that color harmony demonstrates a universal aspect within specific contexts. Generally, certain impressions maintain universal color palettes across diverse settings, while others are context-dependent. There are universally accepted color palettes noted for their considerable aesthetic appeal and several universally perceived color associations. The perception of aesthetics is partly general and partly domain-specific.
\end{itemize}

As we can see from the above, the level of interest in this field appears to be substantial. However, there are also some technical and semantic limitations to research. One important constraint on most works discussed in this area is the limited number of colors and their crisp nature, which is in contrast to human color perception. Some of the experiments were conducted without a validated dataset, and psychological experiments were conducted without having control of participants' conditions.

\subsection{Models of Emotion}

Several emotion models have been proposed to describe and categorize human emotions. Here are a few examples of well-known ones: Russell's Circumplex Model \cite{Russell1980}, Ekman's Six Basic Emotions \cite{ekman1992argument}, Plutchik's Wheel of Emotions, James-Lange Theory \cite{james1884emotion}, Cannon-Bard Theory \cite{cannon1915bodily},\cite{cannon1929emotion}, Barrett's Conceptual Act Theory \cite{barrett2017emotions}, The Geneva Emotion Wheel \cite{scherer2000emotion}, Schachter-Singer Two-Factor Theory \cite{schachter1962cognitive}, Mehrabian's PAD Model \cite{mehrabian1989semantic}.

\begin{figure*}[tb]
\centerline{\includegraphics[width=\textwidth]{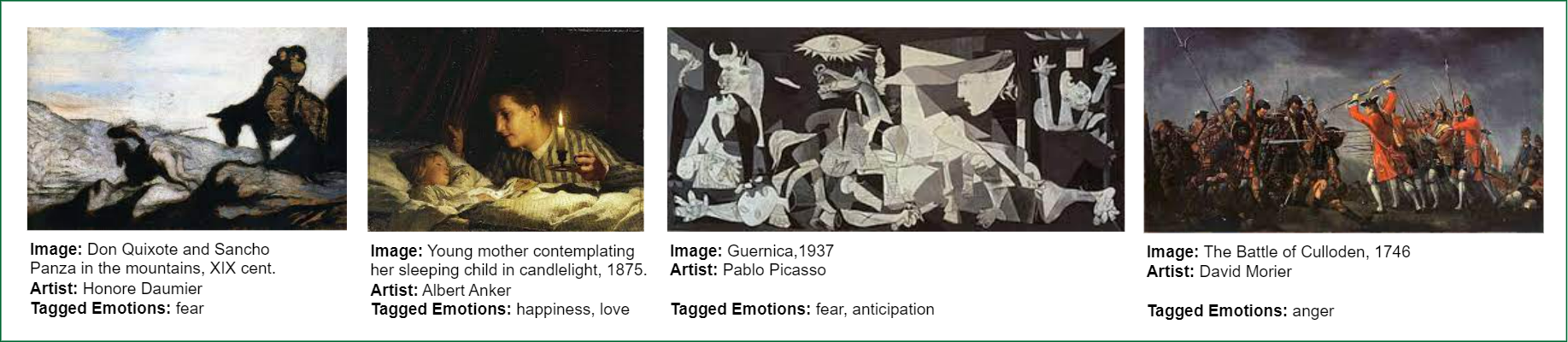}}
\caption{WikiArts Emotions Dataset \cite{WikiArt}. Sample images.}
\label{samples}
\end{figure*}

In our research, we are using Russell's emotional model. Russell's Circumplex Model is a two-dimensional model organizing emotions based on two primary dimensions: valence (pleasure-displeasure axis) and arousal (activation-deactivation axis)\cite{Russell1980}.  According to Russell's approach, emotions don't exist in discrete categories but along continuous dimensions. This is consistent with the fuzzy approach, which allows for the fuzziness and transitionality of emotional borders. Color and emotion relationships in art may be difficult to categorize using rigid boundaries. Thus, any emotion can be, to some extent, pleasant/unpleasant and active/passive.
 
The model was developed based on subjective feelings \cite{Russell1980}. Participants were asked to group 28 emotion words based on perceived similarity, and a statistical technique was used to group related words.

As you can see in Figure \ref{fig:Russell}, a set of emotions is distributed on the wheel. We are using some of them as tags for art paintings.



\begin{table}[t]
\caption{Distribution of emotions in dataset}
\setlength{\arrayrulewidth}{0.5mm}
\setlength{\tabcolsep}{25pt}
\renewcommand{\arraystretch}{1.5}
\begin{tabularx}{0.5\textwidth}{ ccc } 
\hline
\textbf{Emotion} & \textbf{Dataset} & \textbf{Image count} \\
\hline
happiness & 50\% & 773 \\
love & 50\% &  160 \\
anger & 50\% &  21 \\
sadness & 50\% &  159 \\
gratitude & 50\% &  8 \\
fear & 50\% &  202 \\
shame & 50\% &  5 \\
surprise & 50\% &  532 \\
shyness & 30\% &  6 \\
trust & 50\% &  358 \\
\hline
\end{tabularx}
\label{emotion_set}
\end{table}

\section{Methods}

In this paper, we propose a classification system that takes an art piece and predicts the emotional response it can generate.

We aim to enhance the current image emotion classification and retrieval algorithms by utilizing fuzzy logic. Our approach comprises several stages. Firstly, we will identify the most relevant emotions within the art domain and acquire a relevant dataset. We will then proceed to process the images by smoothing and normalizing them. Subsequently, we will retrieve color palettes from the prepared data and map them to image emotion tags. To increase the precision of the mapping process, we will convert the RGB color palettes into fuzzy palettes.

\subsection{Dataset}

To proceed, we must first locate a set of paintings that have been previously tagged with specific emotions. This step can be divided into two parts: training and testing. We use Wiki Art Dataset. This resource contains works from 195 artists, 42129 images for training, and 10628 images for testing \cite{WikiArt}. This dataset has classified 20 emotions into three categories: positive, negative, and other/mixed.

The images in this dataset were manually tagged with the emotions they evoke in humans. At least ten annotators annotated each work of art. There are 3 sets of annotations, namely, based on title, image, and combined (image and title). For our study, we selected annotations based on the image only (without a title) because our analysis is based on colors in the image, ignoring any associated text, including author, title, and genre.

 The labeled dataset is divided into three variants according to aggregation thresholds of 30\%, 40\%, and 50\%, respectively. A label is selected if at least 30\%, 40\%, and 50\% of the responses, or three out of every 10 respondents, suggest that a particular emotion is applicable. The authors claim that 40\% of the dataset is generous aggregation criteria; nonetheless, we selected 50\% of the dataset for nine emotions and 30\% of the dataset for the emotion of \textit{shyness}, which only contains one image in the 50\% of the dataset (see Table \ref{emotion_set}). 
 
 Using this dataset and comparing it with Russell's emotional model, we chose the main emotions for our model. As you can see in Table \ref{emotion_set}, these are the ten most relevant emotional words and their distribution in the dataset. Refer to Table \ref{emotion_set} for details.
 
Sample art images from the WikiArts Emotions Dataset are presented in Fig. \ref{samples}

It already has some annotations. As mentioned before, we can compare these results to determine how much the proposed method corresponds to the accuracy of emotion recognition.

\subsection{Theory of Visual Attention} 

Color perception involves categorizing colors through visual recognition and attentional selection \cite{thesis}, \cite{tva}.

Let us go over the main ideas from a theory of visual attention. Bundesen introduces three types of perceptual categories in his work: color, shape, and location categories \cite{tva}. Our model is primarily based on color analysis. Using Bundesen's notation, colors represent emotions as perceptual categories. 

According to Bundesen's theory of visual attention (TVA), perceptual categorization involves identifying whether an input element $x$ belongs to a certain category $i$. This is expressed as $E(x, i)$, with $E(x, i) \in [0, 1]$ representing the certainty of $x$ belonging to $i$. The set of all perceptual units is denoted by $S$, and the set of all categories is denoted by $R$.

The concept of \textit{saliency} refers to the strength of sensory evidence that suggests an element belongs to a perceptual category. This is denoted by the notation $\eta(x,i)$, where $x$ is the element and $i$ is the perceptual category. Additionally, each category is assigned a pertinence value, denoted by $\pi_i$, representing its importance for a specific task $T$. We can calculate the attentional weight of $x$ using pertinence and salience values:
\begin{equation}
\omega_{x} = \sum\limits_{i\in R}\eta(x,i)\pi_{i}
\label{atw}
\end{equation}

It is important to note that in Equation (\ref{atw}), only categories that have positive pertinence values have an impact on $\omega_{x}$.

When it comes to visual attention, two mechanisms are often used - filtering and pigeonholing. \textit{Filtering} refers to the process of selecting an item, denoted by $x$, from a set of items, denoted by $S$, that belongs to a particular category, denoted by $i$. For example, a client might be searching for a dark and dramatic art object, and filtering can be used to narrow down the art catalog to show only those objects that fit this description. 

On the other hand, \textit{Pigeonholing} refers to the process of identifying a suitable category, denoted by $i$, for a given item, denoted by $x$, that belongs to a set of categories, denoted by $R$. In other words, if we have an item, we can use pigeonholing to determine its category. For example, when a manager of the art gallery collects artists' or experts' judgments on some art object and then categorizes them as being, e.g., \textit{happy, romantic, fearful}, etc.

\begin{figure}[b]
\includegraphics[width=0.45\textwidth]{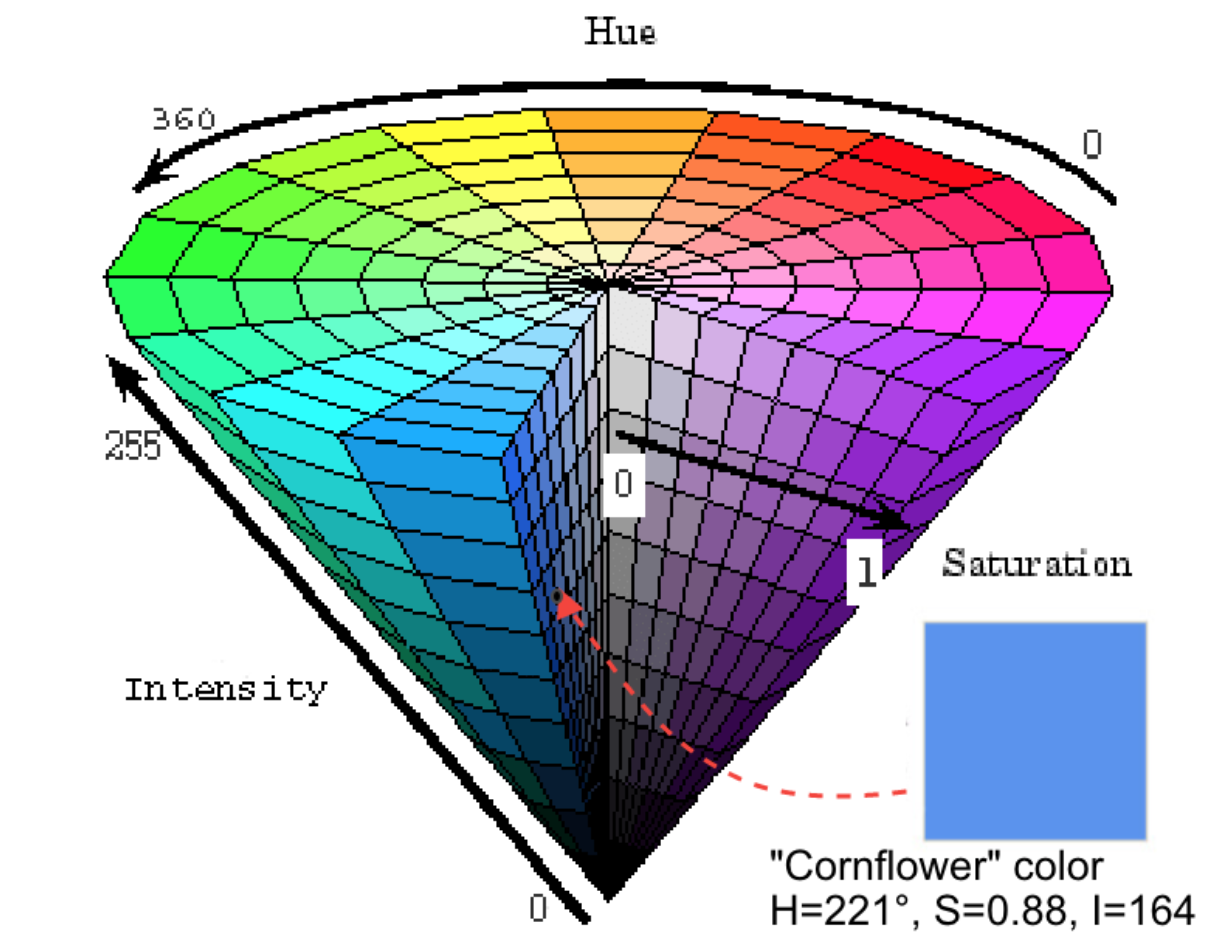}
\caption{Illustration of the HSI color Space \cite{comparative}.}
\label{fighsi}
\end{figure}
\subsection{Fuzzy Sets Theory}

Emotion ambiguity and a few emotion classes compared to human feelings are some difficulties distinguishing emotions \cite{aida}. The fuzzy approach represents a valuable method for dealing with imprecision, and it has been utilized in several research to represent emotions and their intensity in audio, speech, and video \cite{video2023fuzzy, 8957487, 8302926, 9232571}, color image \cite{Hibadullah2015, shamoi2023universal, Shamoi2022, 1527744, Lee2011, visualemotion}, text \cite{peerj, 8722967, 9863839, 31878}, face \cite{7752782, 4505336, 8957487, 8302926} and product \cite{7468477, fss} emotion analysis.

Fuzzy colors are suitable for contextual relevance: they are often associated with specific contexts or cultural domains, greatly influencing emotional responses. By incorporating a fuzzy sets approach, accounting for the cultural, contextual, and individual variations in color-emotion associations leads to a more comprehensive analysis. The next section presents fundamental concepts from fuzzy theory utilized in our study.

\subsubsection{Fuzzy Sets}
Fuzzy sets, first introduced by Zadeh \cite{zadeh}, allow membership degrees to be indicated by a number between 0 and 1. In Boolean logic, we only deal with a pair of numbers, {0,1}. However, when it comes to fuzzy sets, we consider all the numbers within a range of [0,1]. This range is called a \textit{membership function} (MF) and is denoted as $\mu _{A}(x)$, which can be used to represent fuzzy sets.

MFs are mathematical techniques for modeling the meaning of symbols by indicating flexible membership to a set. We can use it to represent uncertain concepts like age, performance, building height, etc. Therefore, MF's key function is to convert a crisp value to a membership level in a fuzzy set. We also need to show triangular and trapezoidal MF here.

\begin{figure*}
  \includegraphics[width=\textwidth]{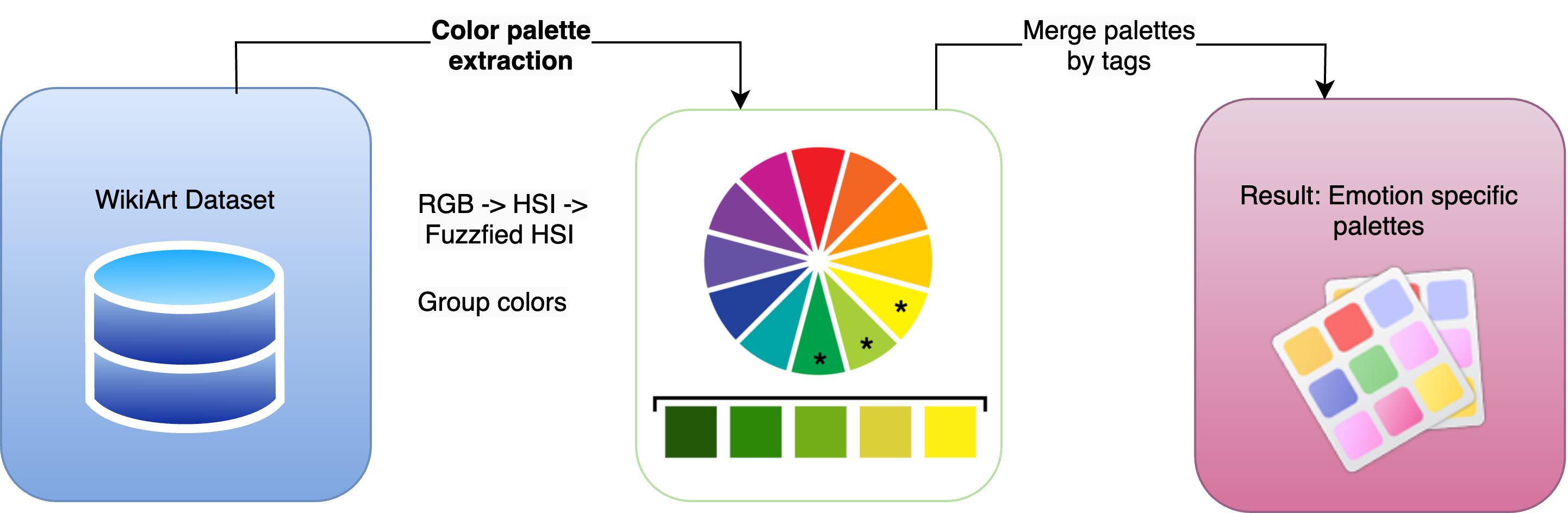}
 \caption{Proposed Approach: Fuzzy colors palettes extraction.}
     \label{fig:fetch palette}
 \end{figure*}

\subsubsection{Linguistic Variables}
Zadeh \cite{Zadeh1975} defined a linguistic variable as a variable whose values are words or sentences in a natural or artificial language rather than numbers. For instance, the label 'deep' can be considered a linguistic value of the variable 'Color Intensity,' similar to a number but with less precision. A linguistic variable's set of all linguistic values is called a\textit{ term set.}

\subsubsection{Fuzzy Hedges}
There are two types of modifiers or hedges: reinforcing and weakening modifiers. These modifiers are used to either strengthen or weaken a statement. The hedge "very" represents the reinforcing modifier:
\begin{equation}
  t_{very}(u) = u^{2}
     \label{veryeq}
\end{equation}

Weakening modifiers are the second type of modifiers.  For example, "more-or-less" is a hedge that indicates a degree of uncertainty.
\begin{equation}
  t_{more\mbox{-}or\mbox{-}less}(u) = \sqrt{u}
     \label{mleq}
\end{equation}
Additionally, the hedge of "not" can be expressed as:
\begin{equation}
  t_{not}(u) = 1-u
     \label{mleq1}
\end{equation}

Hedges can be applied several times. For example, \textit{not very good performance} is the example of a combined hedge consisting of two atomic hedges \textit{not} and \textit{very}.
\subsubsection{Fuzzy Operations}
In fuzzy logic, the $\alpha$-cut (alpha cut) is a crisp set that contains all the members of a given fuzzy subset $f$ whose values are greater than or equal to a certain threshold $\alpha$ (which ranges from 0 to 1). Mathematically, we can represent the $\alpha$-cut of the fuzzy subset $f$ as:

\[
f_{\alpha} = \{x:\mu_{f}(x) \geq \alpha\}
\]

We can use the $\alpha$-cuts to perform set operations on fuzzy sets. For instance, given two fuzzy sets $A$ and $B$, we can obtain their union and intersection by taking the union and intersection of their respective $\alpha$-cuts. That is:

\[
(A \cup B)_{\alpha} = A_{\alpha} \cup B_{\alpha}, \quad (A \cap B)_{\alpha} = A_{\alpha} \cap B_{\alpha}
\]

Fuzzy logic is suitable to cope with subjective metrics, such as emotions and broad concepts, like color because it is consistent with human perception. \cite{Shamoi2022}. Initially, we get the color palette in RGB format; then we translate it into HSI and FHSI.

\subsection{HSI Color Model}

Because of the high correlation between RGB color space attributes and their failure to replicate human color perception, other color spaces, such as HS* (HSL, HSI, HSV), are favored for modeling human color perception \cite{thesis, ojis}.

Three attributes—\textit{hue} (H, such as yellow, cyan, or blue), \textit{intensity} (I, pale vs. dark), and \textit{saturation} (S, saturated vs. dull)—are used in the HSI paradigm to express colors (see Fig. \ref{fighsi}). As a result, it is more in line with how colors are perceived by the human visual system, which is predicated on three interpretations of color: brightness, purity, and category. All three interpretations perfectly correspond to HSI.

The Hue describes the color in terms of an angle between 0 and 360 degrees. The Saturation describes how much the color is mixed with light, with a range between 0 and 100. The Intensity component ranges from 0 to 255, where 0 represents black and 255 represents white.

\begin{table*}[tb]
\caption{Fuzzy attributes of the fuzzy color model. Adapted from \cite{fss}, \cite{thesis}.}
\begin{center}
\centering
\setlength{\arrayrulewidth}{0.5mm}
\setlength{\tabcolsep}{25pt}
\renewcommand{\arraystretch}{1.5}
\begin{tabularx}{\textwidth}{ ccc } 
\hline
\textbf{Fuzzy Variable} & \textbf{Term set } & \textbf{Domain} \\
\hline
Hue            & T = \{ Red, Orange, Yellow, Green, Cyan, Blue, Violet, Magenta \} & X = {[}0, 360{]} \\
Saturation     & T = \{ Low, Medium, High \}                                       & X = {[}0, 100{]} \\
Intensity      & T = \{ Dark, Deep, Medium, Pale, Light \}                         & X = {[}0, 255{]} \\ 
\hline
\end{tabularx}
\end{center}
\label{tab:fhsi}
\end{table*}

\begin{figure*}[htbp]
\centerline{\includegraphics[width=\textwidth]{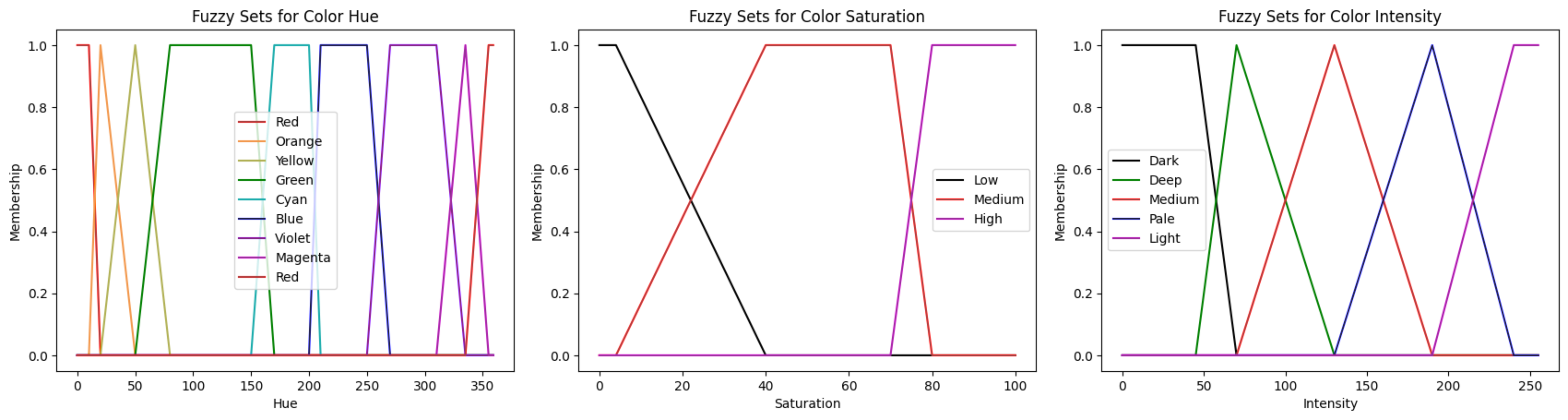}}
\caption{Fuzzy Sets for Hue, Saturation, and Intensity attributes.}
\label{fig:fuzzy_sets}
\end{figure*}

\begin{figure*}
  \includegraphics[width=\textwidth]{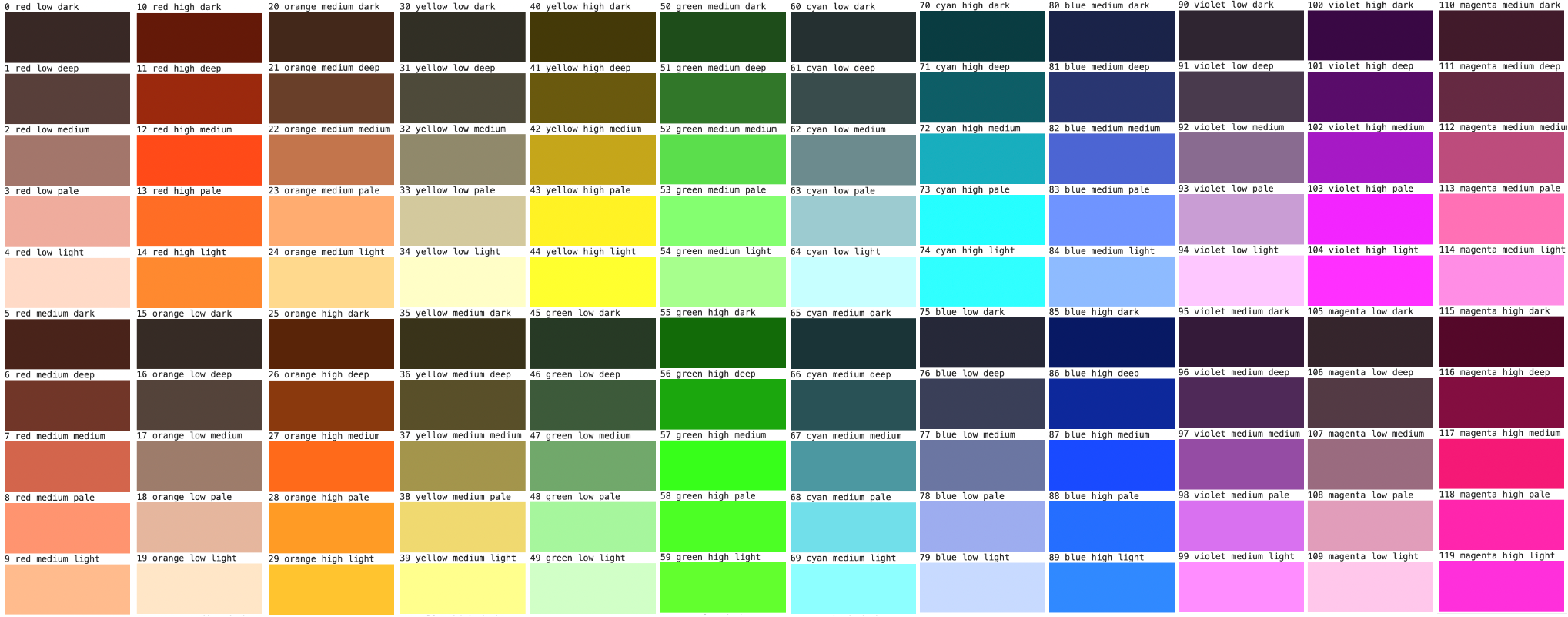}
 \caption{Fuzzy color representation model, which incorporates 120 hues, each characterized by three linguistic terms: hue, saturation, and intensity.}
     \label{fig:many}
 \end{figure*}

\begin{figure*}[bt]
\centerline{\includegraphics[width=\textwidth]{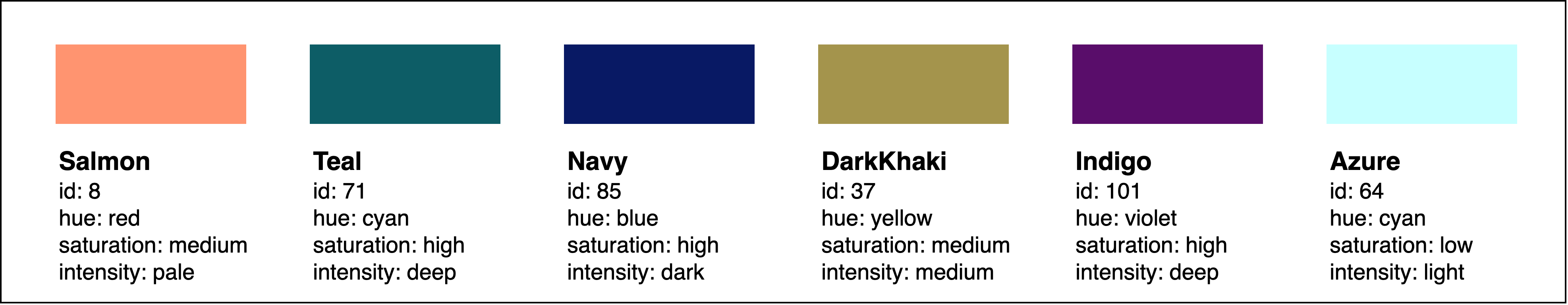}}
\caption{Fuzzy colors examples from the proposed fuzzy color model.}
\label{fig:colors}
\end{figure*}

\subsection{Proposed Approach}
The main idea of the proposed approach can be seen in Fig. \ref{fig:fetch palette}.  First, we process a dataset comprising art images. We extract fuzzy dominant colors and their frequencies associated with certain emotions for each image in the dataset. As a result, we get a fuzzy color distribution for each of the ten emotions. Finally, we convert them back to a crisp domain, obtaining color-emotion associations in the form of basic colors.

We employ a fuzzy approach because it enables us to evaluate emotions in a way that is consistent with human judgment. We obtain a more logical and human-consistent output by partitioning the set of potential emotions into subsets corresponding to linguistic tags \cite{jaciii}. Similarly, fuzzy sets make it simple to show how an emotion gradually changes from one label to another \cite{Zadeh1975}.

In the theory of visual attention, an art image is represented by the variable $x$, while $S$ represents a database of all images. The variable $i$ represents an emotion category, such as fear or happiness, which can be composite (e.g. love and anger). The collection of all emotion categories is represented by $R$.

\subsubsection{Fuzzification of HSI Colors}

Triangular Membership Function:
\[ \mu(x) = \begin{cases}
    0, & x \leq a \\
    \frac{{x - a}}{{b - a}}, & a \leq x \leq b \\
    \frac{{c - x}}{{c - b}}, & b \leq x \leq c \\
    0, & x \geq c
\end{cases} \]

Trapezoidal Membership Function:
\[ \mu(x) = \begin{cases}
    0, & x \leq a \\
    \frac{{x - a}}{{b - a}}, & a \leq x \leq b \\
    1, & b \leq x \leq c \\
    \frac{{d - x}}{{d - c}}, & c \leq x \leq d \\
    0, & x \geq d
\end{cases} \]

Fuzzy partitions were adapted from \cite{fss},\cite{thesis}, \cite{ojis}, \cite{jaciii}, but the changes were done to the "Low" Saturation because we don't have "Any" Category. These partitions were obtained based on a survey on human color categorization.

 Table \ref{tab:fhsi} shows the information about term sets and domains of each fuzzy variable in our color model (H, S, I). Figure \ref{fig:fuzzy_sets} presents the membership functions for fuzzy H, S, I variables. Figure \ref{fig:many} presents the fuzzy colors of the proposed representation model, which incorporates 120 distinct fuzzy colors.  We have divided the intensity into five fuzzy sets, namely "Low," "Medium," and "High," with the domain X = [0, 100] and the universal set U = {0, 1, 2,..., 99, 100}. Similarly, the attribute of hue has been divided into eight fuzzy sets, including "Red," "Orange," "Yellow," "Green," "Cyan," "Blue," "Violet," and "Magenta," with the domain X = [0, 360] and the universal set U = {0, 1, 2,..., 359, 360}. Lastly, the attribute of saturation has been divided into three fuzzy sets, namely "Dark," "Deep," "Medium," "Pale," and "Light," with the domains X = [0, 255], and the respective universal set U = {0, 1, 2,..., 254, 255}. By doing this, we have created a comprehensive and accurate spectrum of fuzzy colors.

 Some color representatives can be seen in Fig \ref{fig:colors}. On the figure, we have a table like Hue: red, Intensity: deep, etc., and an explanation that a fuzzy color is a region, not a point.

A fuzzy color is a subset of points within a crisp color space, which in this case is the HSI space, as mentioned in \cite{soto, thesis, fss}. We define the domains of the attributes H, S, and I as $D_{H}$, $D_{S}$, and $D_{I}$, respectively.
\begin{defn}
\label{def1}
\textit{ \textbf{Fuzzy color} $C$ is a linguistic label whose semantic is represented in crisp HSI color space by a normalized fuzzy subset of $ D_{H} \times  D_{S}  \times D_{I} $.}
\end{defn}

   It can be inferred from Definition~\ref{def1} that there is always at least one crisp color that fully belongs to any given fuzzy color $C$. This prompts us to extend the concept of fuzzy color to the notion of fuzzy color space. As per the fundamental idea of a fuzzy color space \cite{soto}:
    
\begin{defn}
\label{def2}
 \textit{\textbf{Fuzzy color space} is a set of fuzzy colors that define a partition of $ D_{H} \times  D_{S}  \times D_{I} $.}
\end{defn}


\begin{defn}
\label{def3}
\textit{\textbf{ Fuzzy color palette} is a combination of several fuzzy colors \cite{shamoi2023universal}. }
\end{defn}

Each element in a fuzzy color palette is a fuzzy color (area) rather than a sharp color (point) \cite{shamoi2023universal}. Let us take \textit{Salmon} color as an example (see Fig. \ref{fig:colors}). We convert crisp inputs into fuzzy sets for the fuzzification process. For example, if the color is in RGB format (\textit{Salmon}: R=255, G=160, B=122), we first convert it into HSI model (H = 17, S = 32\%, I = 179\%), then to the fuzzy color model (H = \textit{Red}, S = \textit{Medium}, I = \textit{Pale}. Hue, in this case, is partial 'Red' and 'Orange',  while Saturation is partially ‘Medium’ and partially ‘Low’.

\begin{algorithm}
 \KwData{image $ M_{1} $ in RGB format}
 \KwResult{The histogram that represents the most prominent color in image $ M_{1} $ }
 \SetKwProg{Fn}{Function}{}{}
  \tcc{
    The image's dominant color histogram, $ C_{H}(image) $ is a vector $ (h_{C_{1}}, ... , h_{C_{n}} ) $, where each element $ h_{C_{i}} $ represents the frequency of color $ C_{i} $ in the image. }
  \Fn{FindFuzzyDomColors (image)}{
    $FuzzyColors$ $\leftarrow$ an empty dictionary\;
    $FuzzyDomColors$ $\leftarrow$ an empty array\;
    \tcc {initialize the frequency of each fuzzy color to 0}
	 \While{not at end of image}{
          read current pixel\;
          process current pixel\; \tcc{convert a pixel from RGB to HSI, then fuzzify - convert to fuzzy HSI}
          $fc$ $\leftarrow$ computed fuzzy color\;
          $FuzzyColors[fc]++$\;
     }
     \tcc{Return the five most frequent fuzzy colors. The number of colors returned depends on the context of the application.}
      $FuzzyDomColors$ $\leftarrow$ 5 keys from $FuzzyColors$ with max frequency\;
     \Return   $FuzzyDomColors$;
    }
\caption{Finding the image dominant colors \cite{fss, thesis}.}
\label{alg1}
\end{algorithm}

\begin{algorithm}
 \KwData{Dataset of images $ M_{1},...,M_{n} $ of some Emotion $ E $}
 \KwResult{fuzzy colors palette $ FP $ for Emotion $ E $}
 \SetKwProg{Fn}{Function}{}{}
  \tcc{This function calculates the dominant color histogram of the given emotion image, denoted by $ E_{H}(image) $. The resulting output is a vector $(h_{C_{1}}, ..., h_{C_{n}})$ where each $h_{C_{i}}$ element represents the frequency of the fuzzy color $C_{i}$ in the emotion image.}
  \Fn{GetEmotionFuzzyColorsPalette ($emotion$)}{
  $emotionFuzzyColors$ $\leftarrow$ an empty dictionary\;
    $emotionFuzzyDomColors$ $\leftarrow$ an empty array\;
    \tcc {Set the frequency count for each fuzzy color to zero. }
    \For{each $M_i$ in dataset}{
        $FC_i$ $\leftarrow$ FindFuzzyDomColors($M_i$)\;
        \tcc {calculate the frequency of each fuzzy color for emotion. }
        \For{each $fc$ in $FC_i$}{
            $emotionFuzzyColors[fc]++$\;
        }
      }
 \tcc{Return the 15 most frequently occurring fuzzy colors.}
  $emotionFuzzyDomColors$ $\leftarrow$ 15 keys from $emotionFuzzyColors$ with max frequency\;
 \Return   $emotionFuzzyDomColors$;
}
\caption{Receiving emotion fuzzy colors palette. }
\label{alg2}
\end{algorithm}

\begin{figure}[ht]
\includegraphics[width=0.5\textwidth]{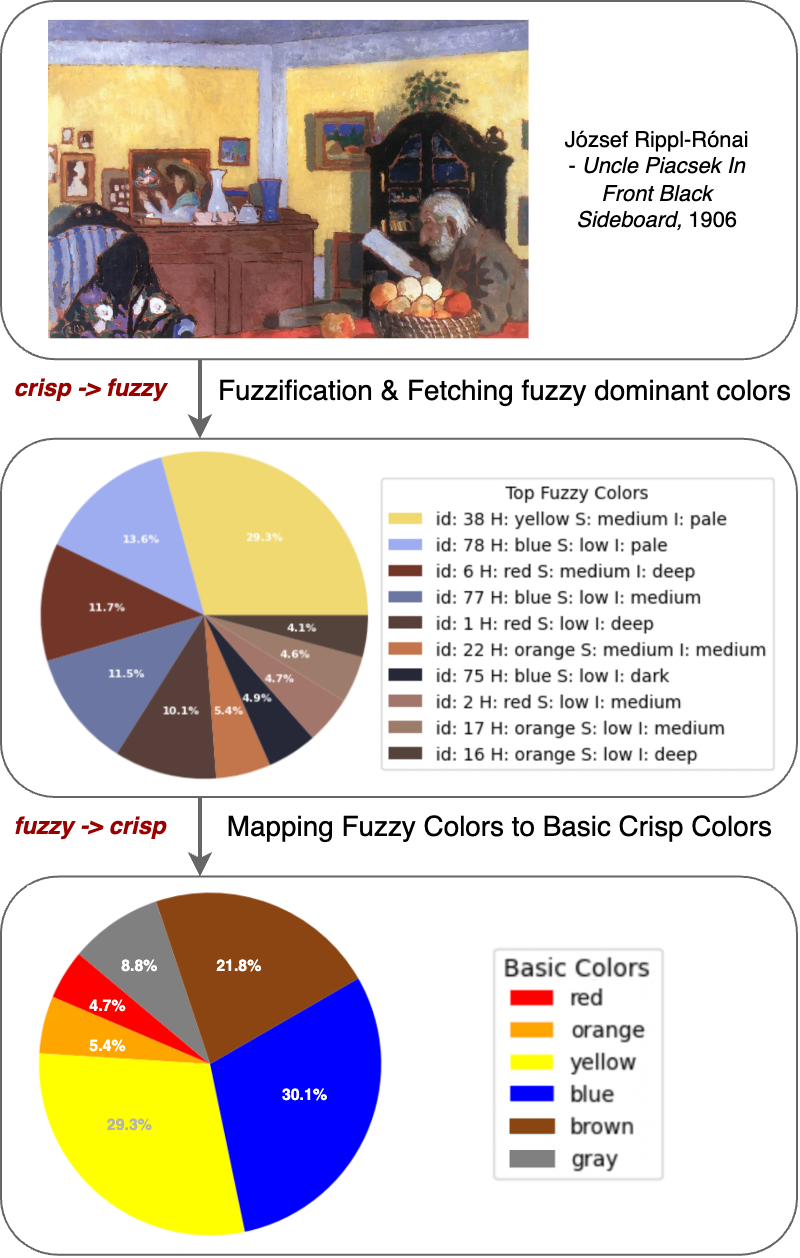}
\caption{Extracting the fuzzy dominant colors from art images and mapping them to basic crisp colors.}
\label{fuzzycrisp}
\end{figure}

\begin{figure*}[ht]
  \includegraphics[width=\textwidth]{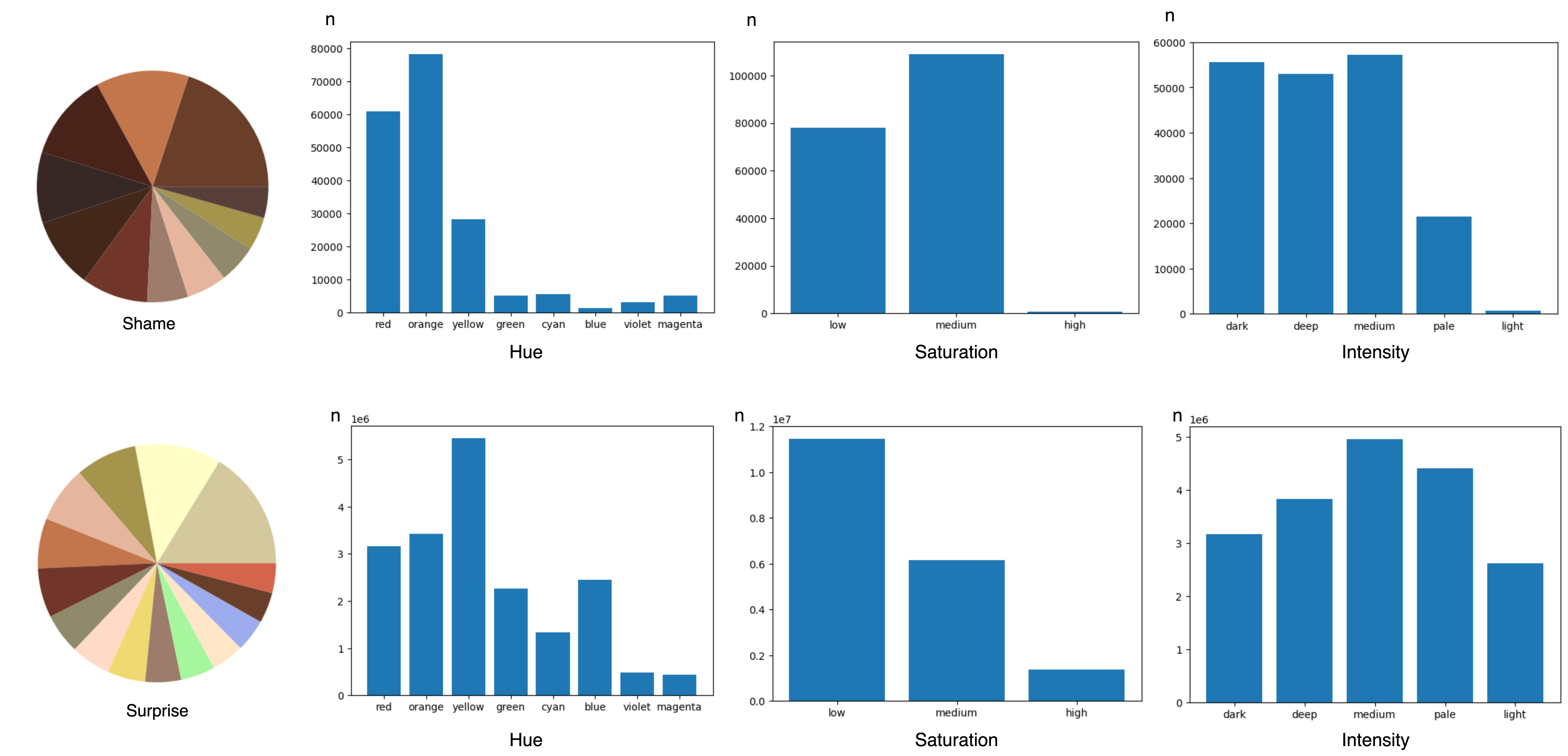}
 \caption{Distribution of HSI attributes across certain emotion}
     \label{fig:barcharts}
 \end{figure*}

\begin{figure}[ht]
    \centering
    \includegraphics[width=1\linewidth]{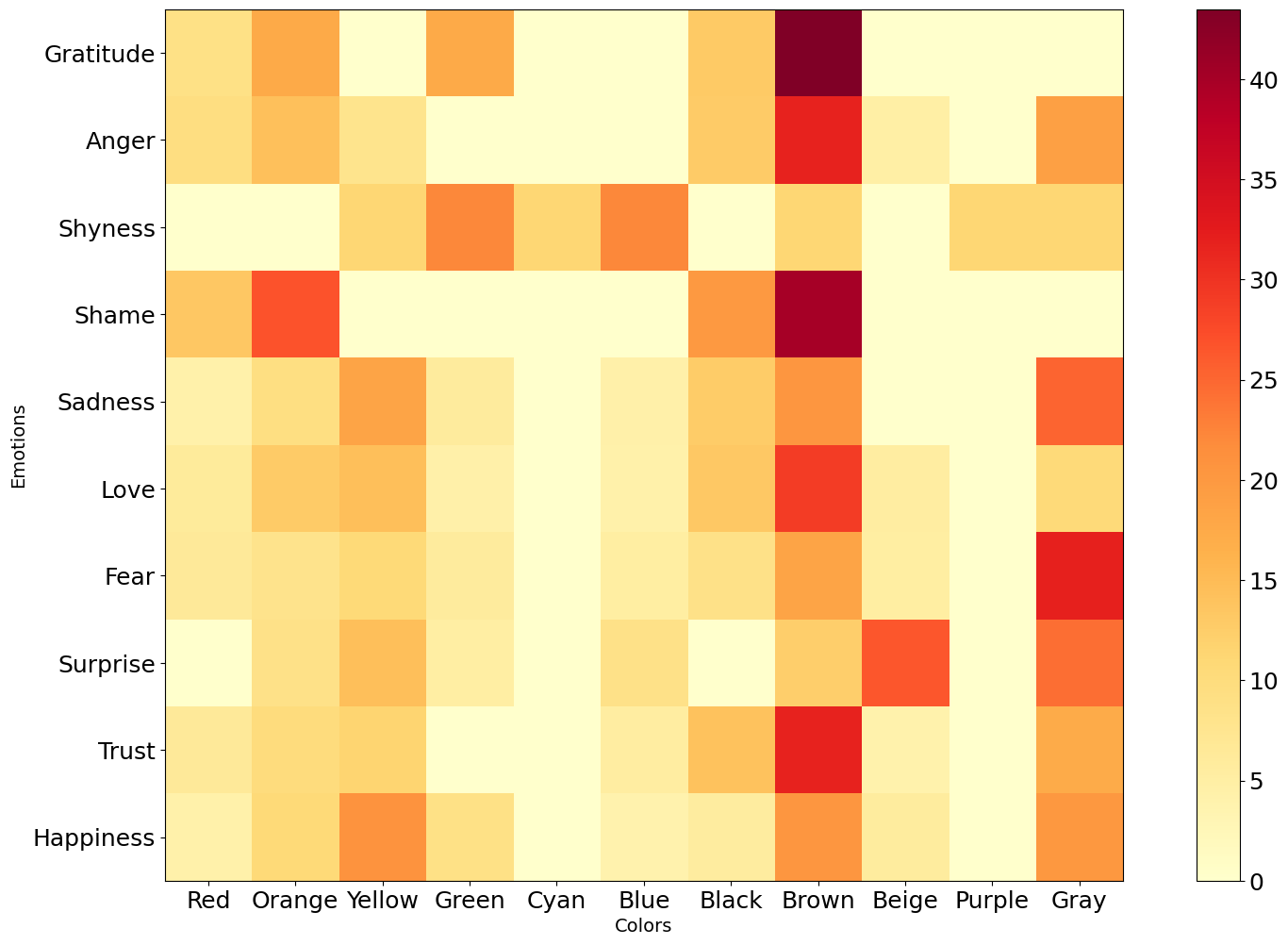}
    \caption{Heatmap of color-emotion association}
    \label{fig:heatmap}
\end{figure}

\subsubsection{Selected Emotions}

This research aims to explore the connection between color palettes and human emotions. As emotions are highly subjective and influenced by various factors such as gender, race, mood, and life experiences \cite{Clarke2008}, we have chosen to focus on the context of art to obtain a more objective assessment. To achieve this, we have examined various online galleries, articles, and datasets to identify which emotions are most commonly associated with paintings. Additionally, Kang et al. have already researched color and emotion pairings, as detailed in their article \cite{Kang2018}.

We have conducted extensive research on this topic and analyzed various psychological articles, including \cite{Fugate2019, Jonauskaite2020}. Our goal was to select a comprehensive list of emotions based on previous studies, which is crucial for comparing our model's accuracy with existing results.

The selected dominant emotions in an art context are gratitude, happiness, anger, love, trust, anger, fear, surprise, sadness, and shame. As a result, we will conduct further research on these emotions. 


 \subsubsection{Image Preprocessing}
The original art images were in various sizes, but have been normalized to (200,200) and converted to RGB mode. Next, we transformed each pixel to HSI and then fuzzy HSI. 
The algorithm for fuzzy dominant color extraction (or fuzzy color histogram) can be seen in Algorithm \ref{alg1}.

\subsubsection{Detection of Emotion Color Palette}
Algorithm \ref{alg1} aims to compute the dominant color palette of an image represented in RGB format. This process involves converting the RGB image to HSI (Hue, Saturation, Intensity) color space, fuzzifying these colors, and then generating a fuzzy dominant color palette based on the frequency of these fuzzy colors.

Algorithm \ref{alg2} is designed to generate a fuzzy color histogram specific to an emotion $E$ based on a labeled dataset of images representing that emotion. For each image $M_i$ in this dataset, the function extracts the dominant fuzzy colors present within that specific image. This extraction process is executed by employing the method \textit{FindFuzzyDomColors}($M_i$), which returns the top 5 most frequent fuzzy colors of this image denoted as $FC_i$.

Next, we increment each color in $ FC_i$'s frequency for the given emotion palette. Following processing each image in the dataset, the algorithm identifies the 15 most frequently occurring fuzzy colors within the emotion dictionary. These top 15 fuzzy colors, determined by their respective frequencies, are considered the dominant representations of this specific emotion. 

Note that some of the obtained fuzzy palettes contain fewer than 15 fuzzy colors. This is because we also considered the proportion of colors; specifically, we filtered out those that constituted less than 3.5\% of the image from the top 15 fuzzy colors.

As an alternative color representation and for a comparative evaluation of the obtained color-emotion associations with the results of other researchers, we mapped fuzzy dominant colors for each emotion to basic crisp colors, akin to defuzzification. The set of basic color categories we employed included \textit{'red', 'orange', 'yellow', 'green', 'cyan', 'blue', 'black', 'brown', 'beige', 'purple', 'gray'}. Fuzzy colors with \textit{'dark'} intensity were aligned with \textit{'black',} while colors with \textit{'low'} saturation are shades of gray, so they were associated with \textit{'gray'} basic color. Considering the contextual relevance and popularity of beige and brown colors in the domain of art paintings, we separated them from fuzzy colors with \textit{'red', 'orange',} and \textit{'yellow'} hues. Fuzzy colors with \textit{'violet'} and \textit{'magenta'} hues were mapped to \textit{'purple'}. The remaining fuzzy colors with hues \textit{'red', 'orange', 'yellow', 'green', 'cyan', 'blue'} were mapped to their respective crisp colors.

 Figure \ref{fuzzycrisp} illustrates the fuzzification process and the subsequent mapping of fuzzy dominant colors to basic crisp colors. The dominant colors are identified using a fuzzy color representation model in the fuzzification phase. These fuzzy colors, indicative of emotional responses, are then systematically mapped to a set of basic crisp colors.

 
\section{Experimental Results}


\begin{figure*}[ht]
    \centering
    \includegraphics[width=\linewidth]{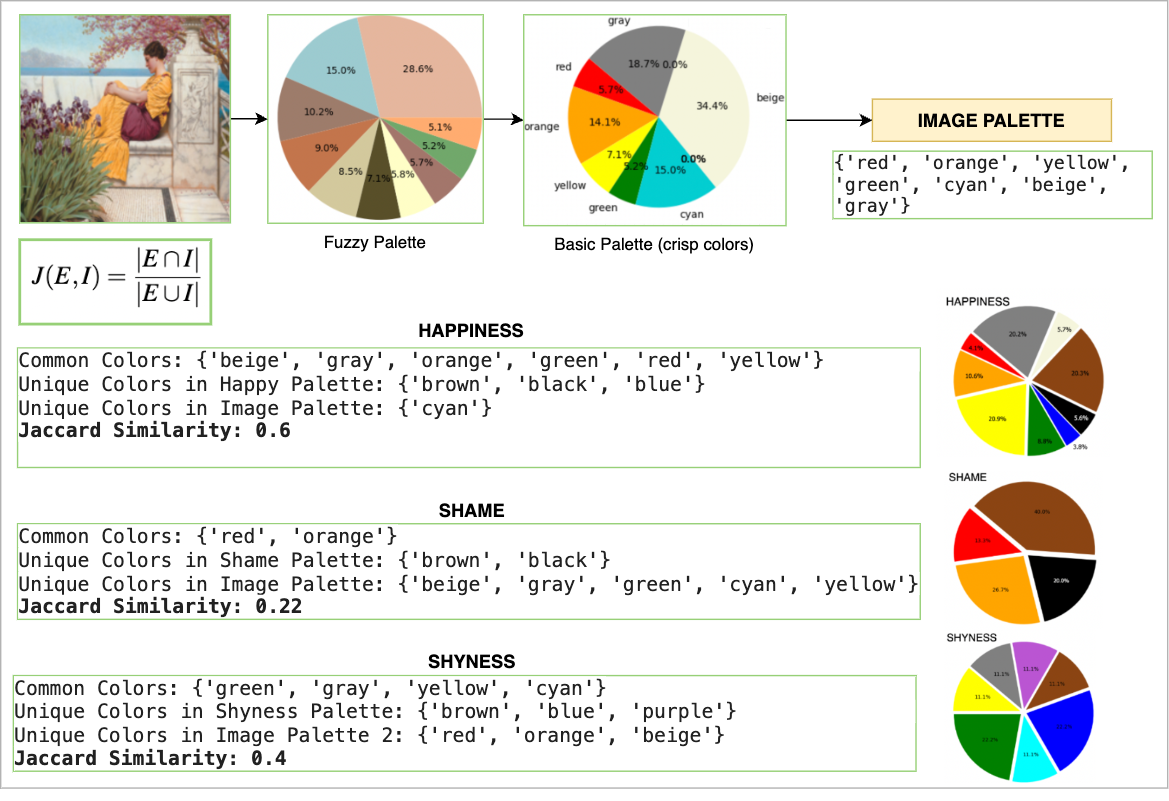}
    \caption{Examples of Jaccard similarity calculation using happy, shy, shame emotion palettes and the John William Godward painting - \textit{Under the Blossom that Hangs on the Bough, 1917.}}
    \label{fig:jaccard}
\end{figure*}

\subsection{Color-emotion Associations Results}
 
        
        Figure \ref{fig:barcharts} illustrates the distribution of hue, saturation, and intensity attributes across certain emotions. One can mention the obvious difference between color palettes for shame and surprise. The two palettes under consideration evoke distinct emotional responses. The first palette, rich in reds, oranges, and yellows, exudes a vibrant and energetic aura with its medium saturation levels and deep, dark intensity. In contrast, the second palette contains a wider range of hues, featuring shades of yellow, green, cyan, and blue, offering a softer appearance due to its lower saturation and a blend of medium to pale intensities. The surprise palette provides cooler tones compared to the intense warmth of the shame palette.

\begin{figure*}[htbp]
  \includegraphics[width=\textwidth]{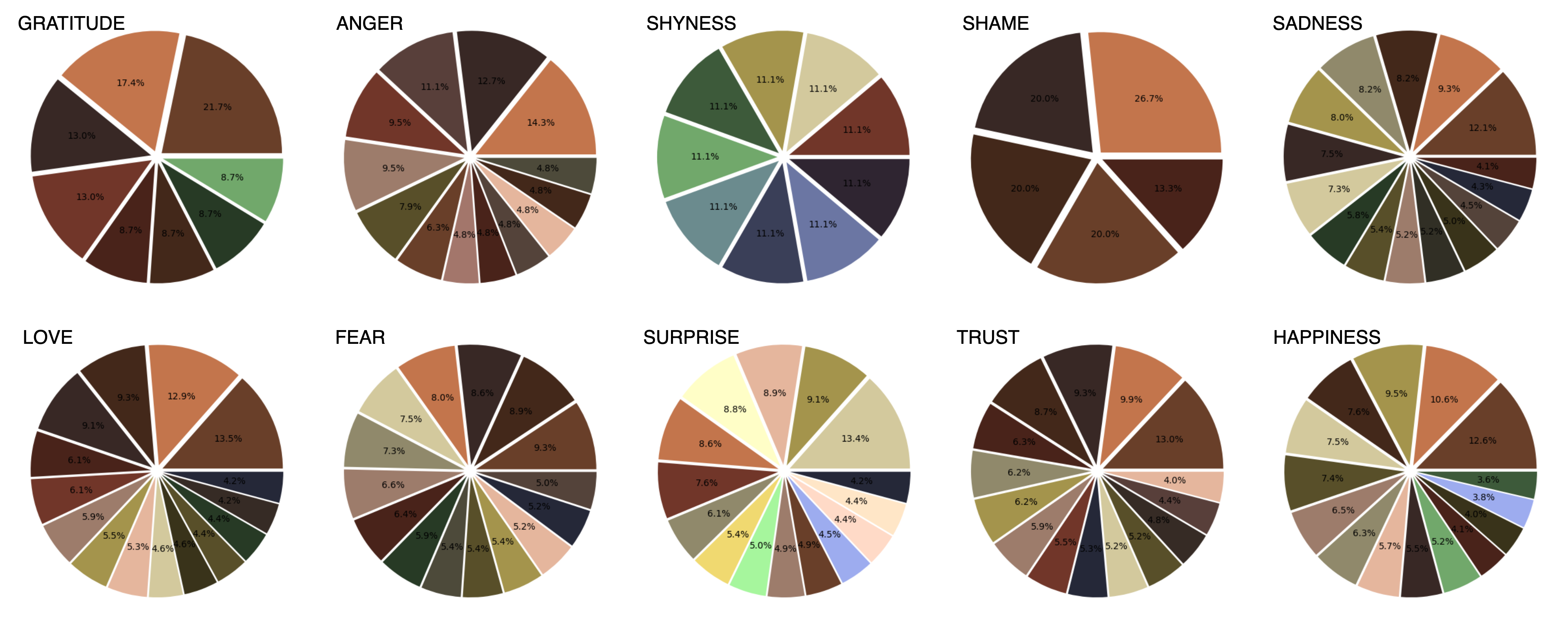}
 \caption{Color-emotion association results: Fuzzy color palettes.}
     \label{fig:fuzzy_results}
 \end{figure*}

\begin{figure*}[htbp]
  \includegraphics[width=\textwidth]{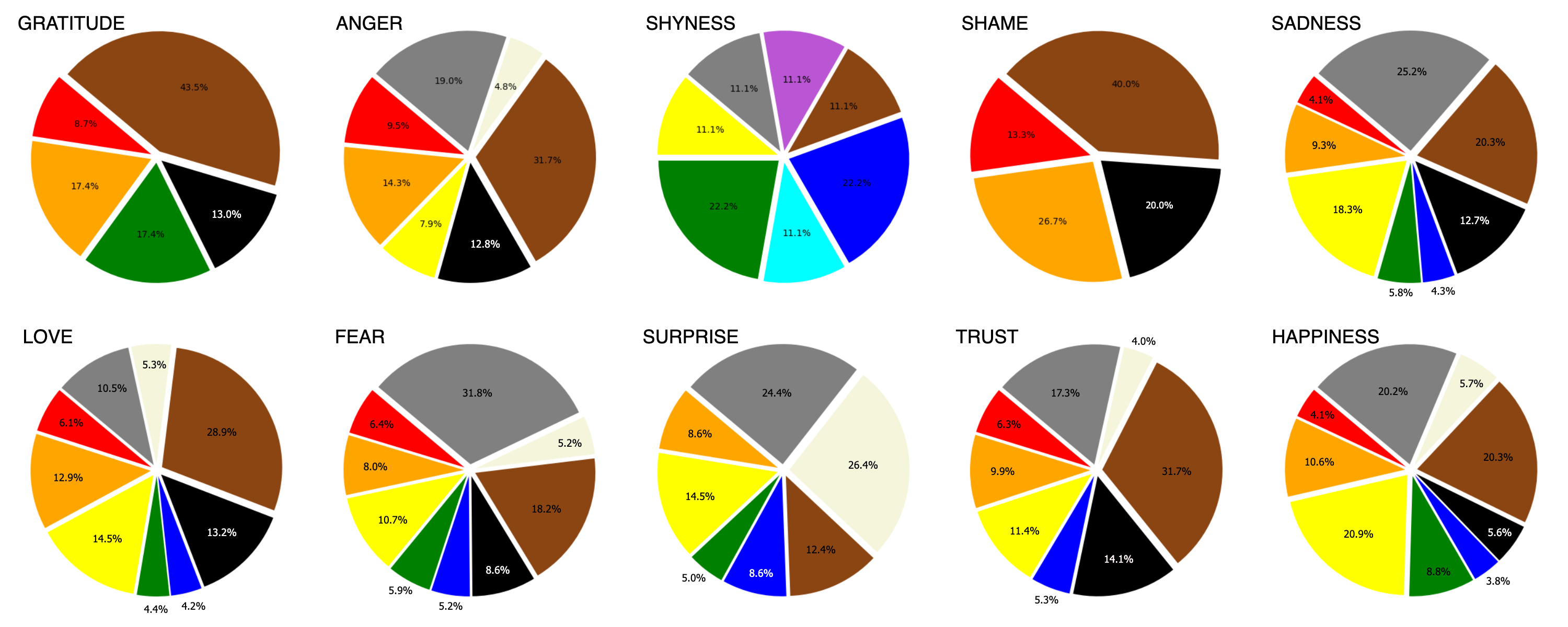}
 \caption{Color-emotion association results: Crisp, basic color palettes.}
     \label{fig:basic}
 \end{figure*}

Our findings suggest that specific emotions exhibit strong associations with certain colors; for example, the '\textit{gratitude'} emotion is strongly associated with \textit{green, brown,} and \textit{orange} colors. Table \ref{tab:basic_colors_results} shows the association of all studied emotions with basic colors. Figures \ref{fig:fuzzy_results}, \ref{fig:basic} provide the visual support for this association.

The associations between emotions and colors reveal interesting patterns. \textit{Happiness} and \textit{love} share similar color associations, prominently featuring brown, green, and orange. \textit{Gratitude} is distinctly linked to brown, green, and orange. \textit{Surprise} stands out as the sole emotion associated with light colors, with an absence of black and a prevalence of beige. \textit{Shyness} is unique, incorporating purple and cyan as basic colors. Shame exhibits a more limited palette, with brown dominating. \textit{Anger} is characterized by brown, gray, black, and red. Gray is notably intertwined with \textit{fear}, and \textit{fear} shares similar color associations with \textit{sadness}. For \textit{trust}, brown and gray take the lead in color associations. Brown consistently appears as the most frequent color in these emotional associations.

 Table \ref{tab:basic_colors_results} illustrates our experimental results. It shows an association between emotions and corresponding basic colors, accompanied by estimated percentages denoting the prevalence or intensity of each color within the representation of those emotions.

Based on our results, prominent color-emotion associations were brown and \textit{gratitude}, brown and \textit{anger}, orange and \textit{shame}, yellow and \textit{happiness}, gray and \textit{fear}. The full spectrum of color associations and emotions can be seen on the heatmap (Fig. \ref{fig:heatmap}).


\begin{table*}
    \caption{Basic colors distribution across emotions (measures indicated in percents)}
    \centering
    \setlength{\arrayrulewidth}{0.5mm}
    \setlength{\tabcolsep}{10pt}
    \renewcommand{\arraystretch}{1.5}
    \begin{tabularx}{\textwidth}{cccccccccccc}
    \hline
    \textbf{Emotion} & \textbf{Red} & \textbf{Orange} & \textbf{Yellow} & \textbf{Green} & \textbf{Cyan} & \textbf{Blue} & \textbf{Black} & \textbf{Brown} & \textbf{Beige} & \textbf{Purple} & \textbf{Gray} \\ [1ex]
    \hline
        Gratitude & 8.7 & 17.4 & 0 & 17.4 & 0 & 0 & 13.0 & 43.5 & 0 & 0 & 0 \\
        Anger & 9.5 & 14.3 & 7.9 & 0 & 0 & 0 & 12.8 & 31.7 & 4.8 & 0 & 19.0 \\
        Shyness & 0 & 0 & 11.1 & 22.2 & 11.1 & 22.2 & 0 & 11.1 & 0 & 11.1 & 11.1\\
        Shame & 13.3 & 26.7 & 0 & 0 & 0 & 0 & 20.0 & 40.0 & 0 & 0 & 0 \\
        Sadness & 4.1 & 9.3 & 18.3 & 5.8 & 0 & 4.3 & 12.7 & 20.3 & 0 & 0 & 25.2 \\
        Love & 6.1 & 12.9 & 14.5 & 4.4 & 0 & 4.2 & 13.2 & 28.9 & 5.3 & 0 & 10.5 \\
        Fear & 6.4 & 8.0 & 10.7 & 5.9 & 0 & 5.2 & 8.6 & 18.2 & 5.2 & 0 & 31.8 \\
        Surprise & 0 & 8.6 & 14.5 & 5.0 & 0 & 8.6 & 0 & 12.4 & 26.4 & 0 & 24.4 \\
        Trust & 6.3 & 9.9 & 11.4 & 0 & 0 & 5.3 & 14.1 & 31.7 & 4.0 & 0 & 17.3 \\
        Happiness & 4.1 & 10.6 & 20.9 & 8.8 & 0 & 3.8 & 5.6 & 20.3 & 5.7 & 0 & 20.2 \\
    \hline
    \end{tabularx}
    \label{tab:basic_colors_results}
\end{table*}

\subsection{Performance evaluation}
In this section, we present the performance evaluation results of the model.

\subsubsection{The Jaccard similarity}
The Jaccard similarity is used to measure the similarity between the two palettes, namely Emotion Palette (E) and Image Palette (I). It calculates the similarity between two sets by comparing their intersection and union. In this case, the intersection is the set of common colors that appear in both palettes, while the union is the total number of unique colors in both palettes.The Jaccard similarity is calculated by dividing the size of the intersection by the size of the union.

The Jaccard similarity value ranges from 0 (completely dissimilar) to 1 (completely similar). A higher Jaccard similarity indicates a greater association between the image and emotion color palettes. Equation (\ref{eq:jaccard}) as follows:
\begin{equation}
J(E, I) = \frac{|E \cap I|}{|E \cup I|}
\label{eq:jaccard}
\end{equation}

Where $|E \cap I|$    is the size of the set of common colors that appear in both image and emotion palettes  $I$ and $E$, $|E \cup I|$ is the total number of unique colors in image and emotion palettes $I$ and $E$.

\subsubsection{Two-alternative Forced choice}
\paragraph{Method description} 
 The experiment involved a total of 177 subjects, bachelor and master students of Kazakh-British Technical University. All subjects who provided answers passed the Ichihara color test. Two students' results were excluded because they did not prove to have normal vision. The experiment utilized a well-known behavioral measure - Two-alternative forced choice (2AFC) \cite{2afc,2afc2}. The primary idea is to compare individuals' real emotion choices based on their evaluations to predicted emotions. 

In one trial of a standard 2AFC experiment, participants indicate which of two visual choices displayed simultaneously they find more appealing or appropriate for a certain category. For every pair that could exist, this process is repeated. Therefore, $\dfrac{n(n-1)}{2}$ trials are needed for 2AFC to measure choices for $n$ stimuli. When a participant is presented with only two mutually exclusive stimuli, it makes the decision job easier for them.
We use The Spearman-Karber method to calculate 2afc measures \cite{2afc}.

\begin{figure}
\includegraphics[width=0.5\textwidth]{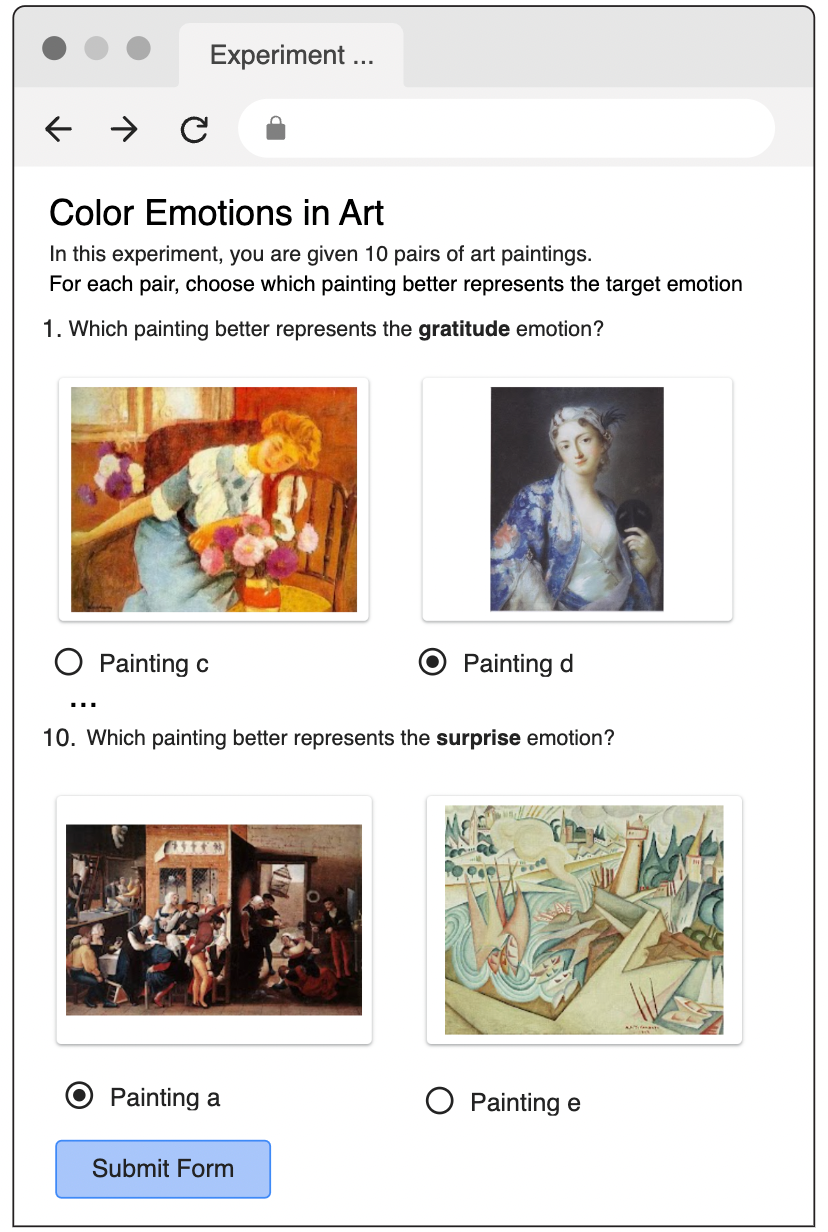}
\caption{2AFC Experiment on perceptual emotion categorization of art paintings. \textit{Art images in the figure: Interior (Lorica), Luchian's last painting; Lady in a Turkish Costume, Felicita Sartori); Brothel scene, Brunswick Monogrammist; Landscape,Amadeo de Souza-Cardoso.}}
\label{fig:experiment}
\end{figure}

\paragraph{Procedure} 
The experiment corresponds to the filtering mechanism of the theory of visual attention \cite{tva}, choosing an item $x \in S $ for a target category $i$.

For each subject, there is one session with 10 questions corresponding to categorization tasks. According to Bundesen's theory of visual attention, perceptual categorization takes the form of "x belongs to i" (represented as E(x, i), where E(x, i) is a value between 0 and 1). In this context, x refers to a perceptual unit and i refers to a perceptual category. The collection of all units is denoted as S, while the collection of all categories is denoted as R. Each question has one perceptual category (e.g., \textit{happy}) and two perceptual units (art paintings).$ S $ contains 5 randomly chosen images. For each predicted emotion, values were calculated. $ R $ corresponds to emotions \textit{\{angry, shyness, happy, sadness, gratitude, shame, fear, trust, love, surprise\}}

On each trial, responders choose art piece that corresponds more to the target emotion. The stimulus pairs differ in predicted emotion intensity. Observers are only allowed to select one item only. After collecting the data, we can compare it to predicted preferences, calculate the "Hits" percentage, and conduct further analyses.

\paragraph{Data Interpretation}  
Equaltion \ref{eqmean} gives the mean of the psychometric function cite{2afc} :
\begin{equation}
  \hat{\mu_{2AFC}} = \frac{1}{2}\sum_{i=1}^{k+1}(\hat{p_{i}}-\hat{p_{i-1}})(x_{i}+x_{i-1})  
   \label{eqmean}
\end{equation}
where $x_{1}< x_{2} ...<x_{k}$ are $k$ monotonically increasing stimulus values, and $\hat{p_{i}} (i=1,...,k)$ are the observed response probabilities,  associated with each stimulus value. To calculate $\hat{\mu_{2AFC}}$, we choose values for $x_0$ and $x_{k+1}$ such that we can assume $p_0=0$ and $p_{k+1}=1$. Note that $x_0$ and $x_{k+1}$ are not part of the experimental design and are only used for the calculation of $\hat{\mu_{2AFC}}$. In our experiment, calculated $ \hat{\mu_{2AFC}} \approx 0.183 $. 

The mean $ \hat{\mu_{2AFC}} $ in 2AFC is also referred to as the point of subjective equality (PSE) \cite{2afc2}. In the context of psychophysics and 2AFC tasks, the PSE represents the stimulus level at which an observer is equally likely to respond "yes" or "no" (in our case, equally likely to choose first or second art image for a target emotion). From Fig. 19 we can see that one of the lowest hit rates correspond to \textit{sadness} and \textit{trust} that have 0.13 and 0.12 stimulus levels, which is less that $PSE \approx 0.183 $.

For the Spearman-Karber method, we need transformed probability values\cite{2afc}. Therefore, the set of observed correct response probabilities $\hat{p}_{i} \ (i = 1, ..., k)$ in a 2AFC can be transformed into the corresponding probability estimates. (see Eq. \eqref{eqpg}).
 \begin{equation}
  \hat{p}_{i} = 2\cdot\hat{g}_{i}-1 
   \label{eqpg}
\end{equation}

\begin{figure}[t]
    \centering
    \includegraphics[width=1\linewidth]{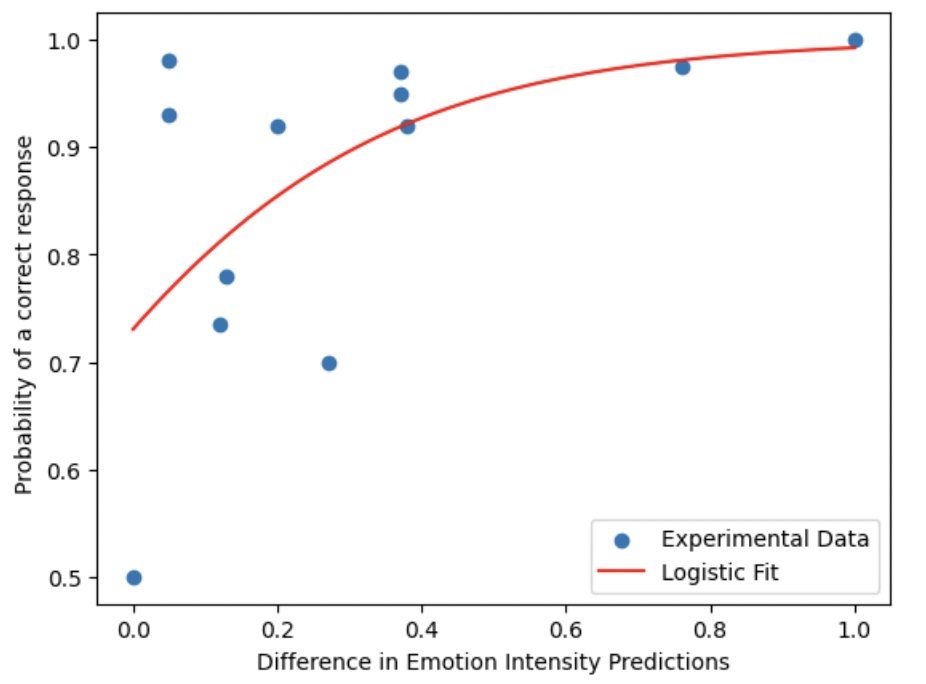}
    \caption{Psychometric function in a choice task. $ \hat{\mu_{2AFC}} \approx 0.183 $, $ SE(\hat{\mu_{2AFC}}) \approx 0.013 $ }
    \label{fig:2afc}
\end{figure}

The experiment's findings are depicted in a psychometric function (Fig. \ref{fig:2afc}), which displays the likelihood of the subject's decision based on the stimulus difference.

 For fitting, the logistic function is used. This figure's abscissa illustrates the difference between two expected emotion intensities. The difference is between 0 and 1. The y-axis shows the proportion of correct responses, ranging from 0.5 to 1.0 based on how distinguishable two stimuli are for the user.

\begin{figure}[tb]
    \centering
    \includegraphics[width=1\linewidth]{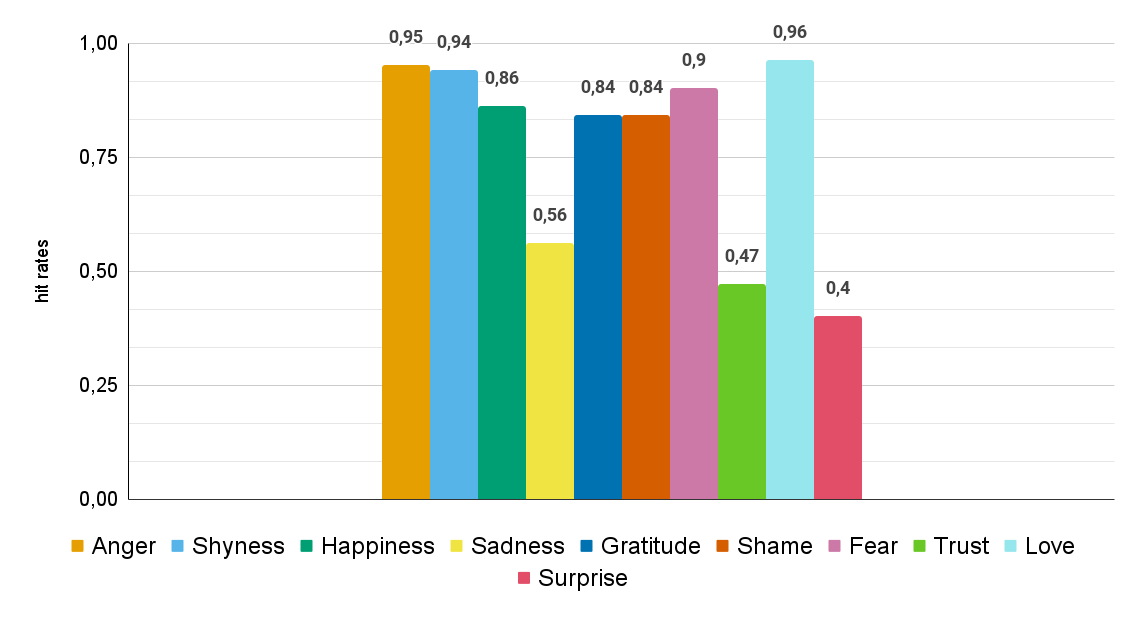}
    \caption{Experiment hit rates distribution across emotions.}
    \label{fig:emotions_hit_rate}
\end{figure}

The average percentage of accurate answers, or hit rate, can be used to evaluate overall performance (see Fig. \ref{fig:emotions_hit_rate}). The average hit rate is 0.77.

\begin{table*}[ht]
    \caption{2AFC experiment results}
    \centering
    \setlength{\arrayrulewidth}{0.5mm}
    \begin{tabular}{ccccccccccc}
    \hline
     & \textbf{Anger} & \textbf{Shyness} & \textbf{Happiness} & \textbf{Sadness} & \textbf{Gratitude} & \textbf{Shame} & \textbf{Fear} & \textbf{Trust} & \textbf{Love} & \textbf{Surprise} \\
    \hline
         \# hit rates & 165 & 163 & 148 & 97 & 146 & 146 & 156 & 81 & 166 & 70 \\
         hit rate, \% & 0,95 & 0,94	& 0,86 & 0,56 &	0,84 &	0,84 &	0,9	& 0,47 & 0,96 & 0,4 \\
         Difference in Emotion Predictions & 0,76 & 0,37 & 0,05 & 0,13 & 0,2 & 0,38 & 0,37 & 0,12 & 0,05 & 0,27 \\
    \hline
    \end{tabular}
    \label{tab:2afc_results}
\end{table*}

Next, let us find the standard error, which is also referred to as a "perceptual noise" \cite{2afc} (see Eq. \eqref{eqse}):
 \begin{equation}
  SE(\hat{\mu_{2AFC}}) = \sqrt{\sum_{i=1}^{k}\dfrac{\hat{g}_{i}\cdot(1-\hat{g}_{i})}{n_{i}-1}\cdot  (x_{i+1}-x_{i-1})^{2}}
   \label{eqse}
\end{equation}
where $n_i$ represents the number of observations at stimulus level $i$. The standard error associated with the threshold estimates $\hat{\mu}_{2AFC}$ is approximately equal to 0.013.

\begin{figure}[ht]
    \centering
    \includegraphics[width=1\linewidth]{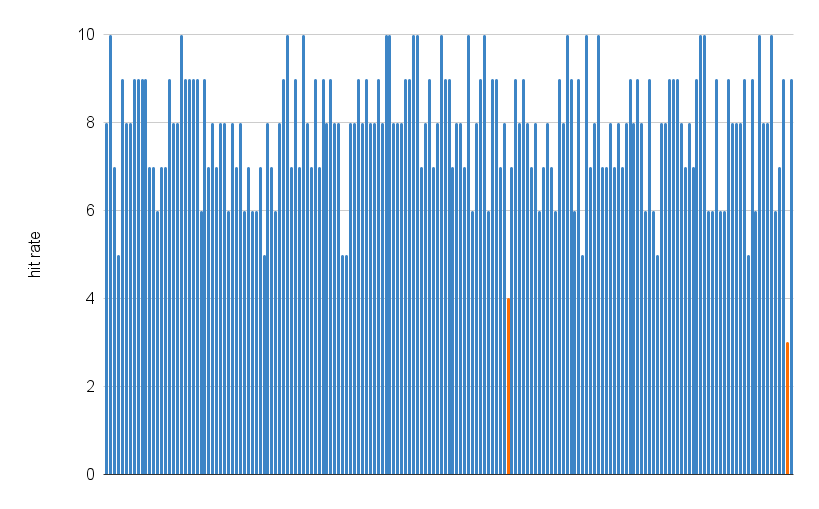}
    \caption{Survey participants hit rates distribution. Outliers are marked with orange color. Their responses were excluded from the analysis.}
    \label{fig:participant_hits}
\end{figure}



As can be seen from Fig. \ref{fig:participant_hits}, we compared participants' responses regarding the most voted answers and highlighted some outliers. According to categorical data analysis, we consider the respondent an outlier in case of less than 50\% responses matching other participants' most selected choice. There are two outliers identified. We also did not analyze the answers of two respondents who did not pass the Ichihara color blindness test (with IDs 107, and 155).

Figure \ref{fig:emotions_hit_rate} illustrates the hit rate for each emotion, namely the proportion of participants that could identify an emotion in images representing that particular emotion according to our algorithm.

Table \ref{tab:2afc_results} represents two alternative forced choice experiment results analysis. The hit rate refers to the number of participants correctly identifying an emotion in line with our model's classification. We also calculate their percentage of total respondents. The difference in emotion intensity predictions signifies the numerical variance between the levels of a particular emotion found in two different images.


\subsection{Sample Application (Proof of Concept)}

The emotion classification method we proposed is simple to adapt for emotion-based retrieval of art paintings. It may be advantageous to retrieve art images from big multimedia archives. The method can be easily adapted for the matching engine that retrieves art objects with similar emotions. For the arbitrary art image, we get the fuzzy color dominant palette (Algorithm \ref{alg1}). Next, we find the relevant emotions to a given art object using the color knowledge base we previously created (See Fig. \ref{fig:fetch emotions}).

\begin{figure}[t]
\includegraphics[width=0.5\textwidth]{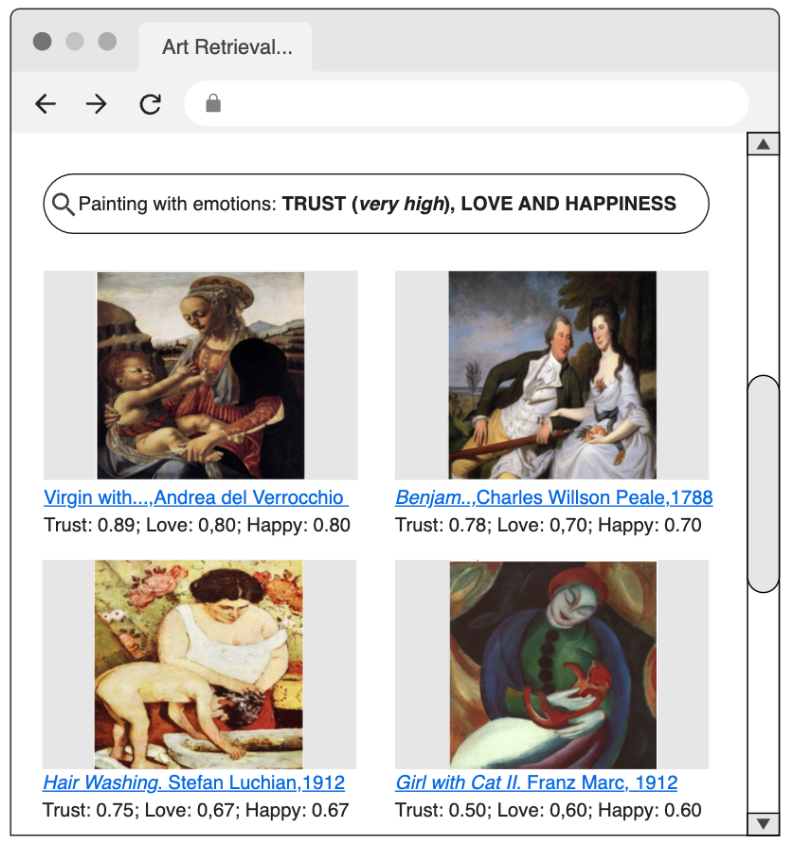}
\caption{Prototype Art Painting Retrieval System}
\label{artsystem}
\end{figure}

\begin{figure*}[htbp]
  \includegraphics[width=\textwidth]{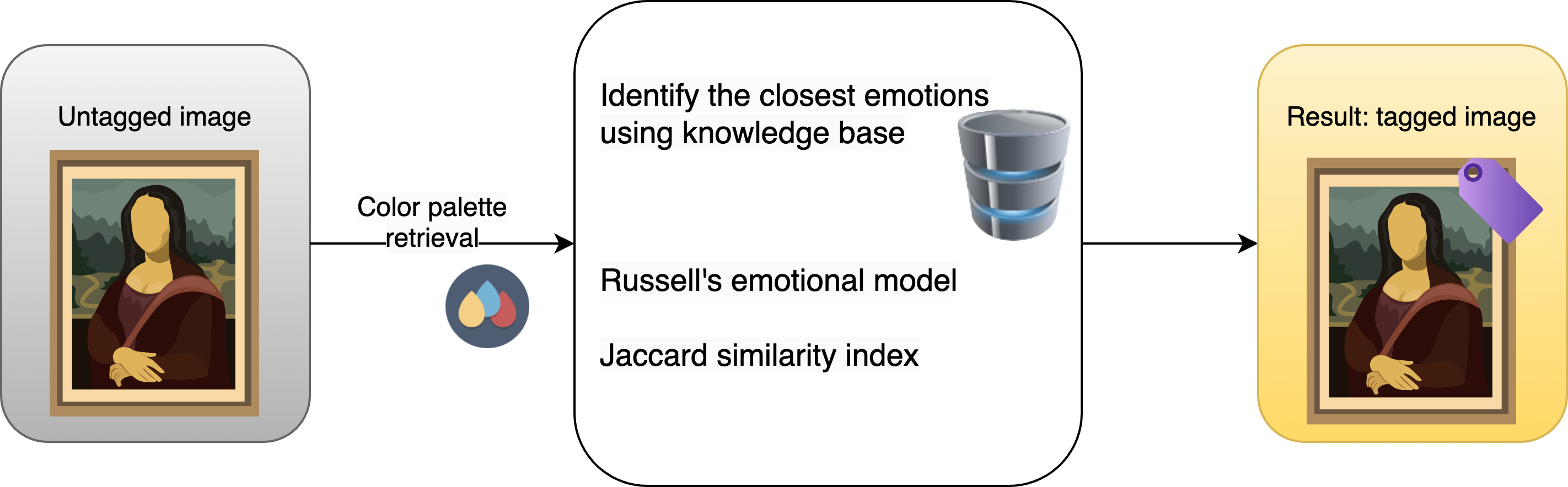}
\caption{Fetching emotions from art image}
\label{fig:fetch emotions}
\end{figure*}

Figure \ref{artsystem} demonstrates the interface of a prototype system for this task. It can potentially use fuzzy natural queries, retrieve emotion labels (e.g., trust) and their desired intensity (e.g., very high), and use fuzzy sets and fuzzy hedges to fetch the corresponding items. The proposed approach can be used in various applications, e.g., in the matching engine, that uses emotions similarity measurement to implement retrieval.

Our method manages uncertainty and imprecision in data. Thus, the system has the potential to assist in recognizing emotions in content, suggesting recommendations, and improving user interactions on interactive platforms.

\section{Discussion}

Several recent studies have explored the emotional effects of art objects on humans \cite{arten}, \cite{Hibadullah2015}, \cite{access_em}, \cite{access_em3}. 
Conceptually similar works were completed by psychologists via surveys and case studies among real people. We identified certain patterns and differences across various emotions by comparing the experimental findings and established psychological studies concerning color-emotion associations. 

Yellow emerged as a dominant color associated with \textit{happiness} in both the experimental findings and established psychological studies\cite{Fugate2019, Jonauskaite2020}
Several studies \cite{Fugate2019, Damiano2023, Domicele2020} matched our results regarding \textit{anger}, identifying it with red and black colors. \textit{Fear} is expressed with gray \cite{Damiano2023} and shame with red  at most researches\cite{Jonauskaite2020, Domicele2020}.
Generally, green colors appeared to be associated mostly with positive emotions, and \textit{shyness} in the experiment was displayed with blue and cyan as dominant colors. Moreover, both sets of studies showed a strong correlation between \textit{sadness} and gray and black.
The results of \cite{Fugate2019} also matched our results in associating green with gratitude, although the experimental findings also noted a similarity with brown.

Conversely, our findings did not align with psychological results regarding \textit{love}-associated colors, with no dominant color evident in this emotional context. Also, the \textit{surprise} primarily featured light tones like beige and yellow, deviating from the typical association with black in psychological studies. 

Common colors such as gray, brown, and black appeared dominantly in both sets of studies. However, blue did not dominate in the experimental findings as it did in psychological studies, highlighting a slight disparity.

Authors of \cite{Hibadullah2015} also used a fuzzy approach to find a relationship between colors and emotion. We can observe specific universal rules governing certain emotions when comparing our findings with their results. For example, \textit{sadness} is represented mostly with black and brown, and blue color is relevant to fear.
Similar results were also received in study \cite{access_em}, where the authors proposed a machine learning model. One of the most significant findings is that \textit{happiness} is undoubtedly associated with yellow and orange, as indicated in both the model prediction outcomes and psychological studies. Also, in this study, \textit{sadness} was correlated with gray color, which is dominant for our \textit{sadness} result palette too.
Generally, \textit{anger} is expressed with red, black, brown, and gray in all the mentioned studies, psychological experiments, and our results.


\section{Conclusion}
In this study, we introduced a novel approach for emotion retrieval using color-based features and fuzzy sets, providing valuable insights into the intersection of computational methods and human emotional perception in visual art analysis.

The system processed a diverse art dataset to generate color palettes associated with ten distinct emotions. We experimentally validated our approach with 2AFC experiment and discovered that it has good predictive power of the color-emotion associations (average hit rate is 0.77).
This suggests a substantial alignment between the emotions inferred by the system and those perceived by human observers, validating the system's ability to approximate emotional states accurately.

Adopting the Fuzzy Sets and Logic approach is advantageous due to its consistency with human perception. Our method enhances algorithm interpretability and adaptability in human-like applications. This system could aid in content-based emotion recognition, recommendation systems, and enhancing user experiences in interactive platforms.

Our research contributes to a more comprehensive understanding of color-emotion associations, offering valuable insights for various practical applications besides art, like marketing, design, and psychology



As for the limitations, the system's performance may vary across different art styles, cultural contexts, and individual perceptions. Next, detecting the closest emotion to an image based solely on its presence doesn't account for varying color frequencies.  Colors, texture, and composition can evoke emotions \cite{new_texture}, creating additional dependencies. We concentrated on color features only, ignoring the objects in the art image, although they also contribute to the overall emotional association. Color can have an impact not only if it is abundant but also if it contrasts sharply with the main color scheme of the background.

Future research could focus on expanding the dataset to encompass more diverse artwork, incorporating contextual information, and refining the fuzzy logic model to enhance accuracy. We see great potential for personalized color-emotion applications. Understanding the emotional nuances of various colors leads to developing personalized color schemes that evoke specific emotional responses tailored to individual preferences or target audiences. We also plan to improve the model with additional dependencies, such as pixel importance.

\bibliography{library}
\begin{IEEEbiography}[{\includegraphics[width=1in,height=1.25in,clip,keepaspectratio]{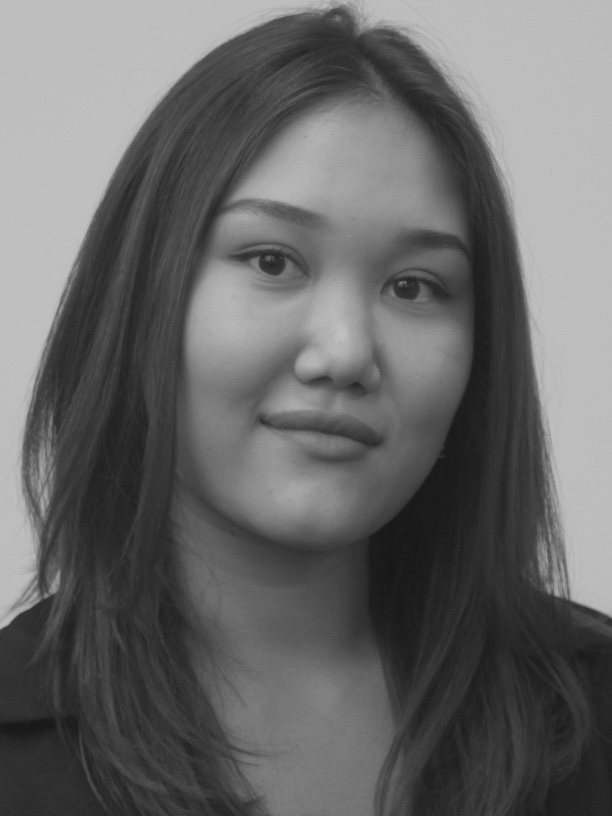}}]{Muragul Muratbekova} received a B.S. degree in information systems from
the Kazakh-British Technical University, Almaty, Kazakhstan, in 2022.  She is currently pursuing an M.S. degree in IT management at the same university and works as a senior software developer in a leading telecommunication company in Kazakhstan. She participated in a number of conferences (EUSPN 2023, FSDM 2023) and one grant project. Her research interests include visual perception computing, emotion engineering, image processing.

\end{IEEEbiography}

\begin{IEEEbiography}[{\includegraphics[width=1in,height=1.25in,clip,keepaspectratio]{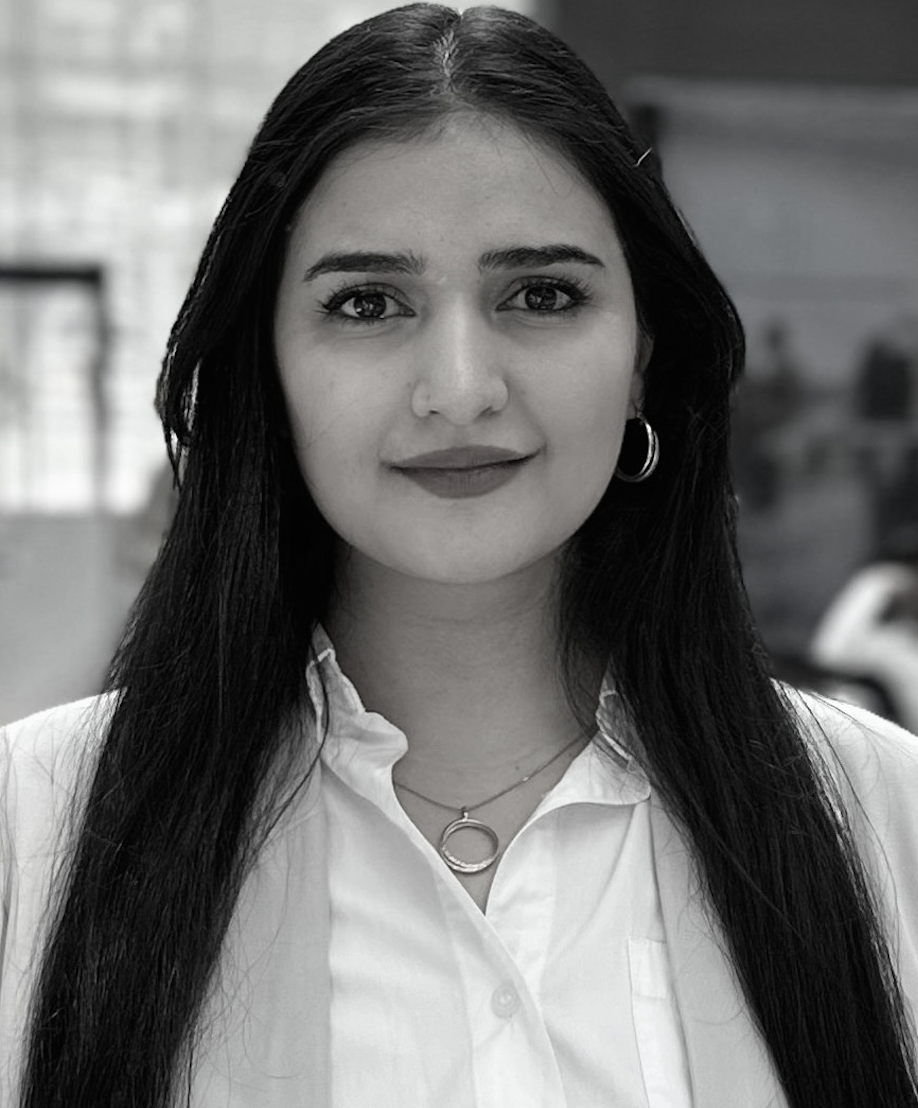}}]{Pakizar Shamoi}  received the B.S. and M.S. degrees in information systems from the Kazakh-British Technical University, Almaty, Kazakhstan in 2011 and 2013, and the Ph.D. degree in engineering from Mie University, Tsu, Japan, in 2019. In her academic journey, she has held various teaching and research positions at Kazakh-British Technical University, where she has been serving as a professor in the School of Information Technology and Engineering since August 2020. She is the author of 1 book and more than 28 scientific publications. Awards for the best paper at conferences were received five times. Her research interests include artificial intelligence and machine learning in general, focusing on fuzzy sets and logic, soft computing, representing and processing colors in computer systems, natural language processing, computational aesthetics, and human-friendly computing and systems. She took part in the organization and worked in the org. committee (as head of the session and responsible for special sessions) of several international conferences - IFSA-SCIS 2017, Otsu, Japan; SCIS-ISIS 2022, Mie, Japan; EUSPN 2023, Almaty, Kazakhstan. She served as a reviewer at several international conferences, including IEEE:
SIST 2023, SMC 2022, SCIS-ISIS 2022, SMC 2020, ICIEV-IVPR 2019, ICIEV-IVPR 2018.

\end{IEEEbiography}


\end{document}